%% file: arxiv.tex
\documentclass[12pt]{article}

\usepackage{graphicx}
\usepackage{amssymb}
\usepackage{amsmath}
\usepackage{amsbsy}
\usepackage{latexsym}
\usepackage{subfigure}
\usepackage{bm}
\usepackage{url}

\RequirePackage{natbib}
\RequirePackage{graphicx}
\bibpunct{(}{)}{;}{a}{,}{,}

 \usepackage{multirow}
 \usepackage{lscape}

\input{mathdef.tex}

\begin{document}

\title{Information-Maximization Clustering\\
  based on Squared-Loss Mutual Information}
\date{}
\author{Masashi Sugiyama (sugi@cs.titech.ac.jp)\\
Makoto Yamada (yamada@sg.cs.titech.ac.jp)\\
Manabu Kimura (kimura@sg.cs.titech.ac.jp)\\
Hirotaka Hachiya (hachiya@sg.cs.titech.ac.jp)\\
Department of Computer Science,
Tokyo Institute of Technology\\
2-12-1 O-okayama, Meguro-ku, Tokyo 152-8552, Japan.
}

\sloppy
\maketitle
\thispagestyle{myheadings}
\markright{}

  \begin{abstract}\noindent
\emph{Information-maximization clustering} learns a probabilistic classifier
in an unsupervised manner so that mutual information
between feature vectors and cluster assignments is maximized.
A notable advantage of this approach is that it only involves continuous optimization
of model parameters,
which is substantially easier to solve than discrete optimization of cluster assignments.
However, existing methods still involve non-convex optimization problems,
and therefore finding a good local optimal solution is not straightforward in practice.
In this paper, we propose an alternative information-maximization clustering method
based on a \emph{squared-loss} variant of mutual information.
This novel approach gives a clustering solution \emph{analytically}
in a computationally efficient way via kernel eigenvalue decomposition.
Furthermore, we provide a practical model selection procedure
that allows us to objectively optimize tuning parameters included in the kernel function.
Through experiments, we demonstrate the usefulness of the proposed approach.

\begin{center}
  \textbf{Keywords}
\end{center}
Clustering, Information Maximization, Squared-Loss Mutual Information.
\end{abstract}

\pagestyle{myheadings}
\markright{}

\section{Introduction}\label{sec:introduction}

The goal of \emph{clustering} is to classify data samples into disjoint groups
in an unsupervised manner.
\emph{K-means} \citep{BerkeleySymp:MacQueen:1967}
is a classic but still popular clustering algorithm.
However, since k-means only produces linearly separated clusters,
its usefulness is rather limited in practice.

To cope with this problem, various non-linear clustering methods
have been developed.
\emph{Kernel k-means} \citep{IEEE-TNN:Girolami:2002}
performs k-means in a feature space induced by a reproducing kernel function
\citep{book:Schoelkopf+Smola:2002}.
\emph{Spectral clustering} \citep{IEEE-PAMI:Shi+Malik:2000,nips02-AA35}
first unfolds non-linear data manifolds by a spectral embedding method,
and then performs k-means in the embedded space.
\emph{Blurring mean-shift} \citep{IEEE-IT:Fukunaga+Hostetler:1975,ICML:Carreira-Perpinan:2006}
uses a non-parametric kernel density estimator for
modeling the data-generating probability density,
and finds clusters based on the modes of the estimated density.
\emph{Discriminative clustering} learns a discriminative classifier
for separating clusters, where class labels are also 
treated as parameters to be optimized \citep{NIPS2005_834,NIPS2007_870}.
\emph{Dependence-maximization clustering} determines cluster assignments so that
their dependence on input data is maximized
\citep{ICML:Song+etal:2007b,ICML:Faivishevsky+Goldberger:2010}.

These non-linear clustering techniques would be capable of handling
highly complex real-world data.
However, they suffer from lack of objective model selection strategies\footnote{
`Model selection' in this paper refers to
the choice of tuning parameters in kernel functions or similarity measures,
not the choice of the number of clusters.
}.
More specifically, the above non-linear clustering methods
contain tuning parameters
such as the width of Gaussian functions and the number of nearest neighbors
in kernel functions or similarity measures,
and these tuning parameter values need to be manually determined
in an unsupervised manner.
The problem of learning similarities/kernels was addressed in
earlier works
\citep{MeiShi01,ICCV:Shental+etal:2003,AISTATS:Cour+etal:2005,JMLR:Bach+Jordan:2006},
but they considered supervised setups, i.e., labeled samples are assumed to be given.
\citet{NIPS17:Zelnik-Manor+Perona:2005}
provided a useful unsupervised heuristic to determine the similarity
in a data-dependent way.
However, it still requires the number of nearest neighbors to be determined manually
(although the magic number `7' was shown to work well in their experiments).

Another line of clustering framework called
\emph{information-maximization clustering}
exhibited the state-of-the-art performance
\citep{NIPS2005_569,NIPS2010_0457}.
In this information-maximization approach, probabilistic classifiers
such as a kernelized Gaussian classifier \citep{NIPS2005_569}
and a kernel logistic regression classifier \citep{NIPS2010_0457}
are learned so that
\emph{mutual information} (MI) between feature vectors and cluster assignments is maximized
in an unsupervised manner.
A notable advantage of this approach is that
classifier training is formulated as
continuous optimization problems,
which are substantially simpler than discrete optimization of cluster assignments.
Indeed, classifier training can be carried out
in computationally efficient manners
by a gradient method \citep{NIPS2005_569}
or a quasi-Newton method \citep{NIPS2010_0457}.
Furthermore, \citet{NIPS2005_569} provided a model selection strategy
based on the information-maximization principle.
Thus, kernel parameters can be systematically optimized
in an unsupervised way.


However, in the above MI-based clustering approach,
the optimization problems are non-convex,
and finding a good local optimal solution is not straightforward in practice.
The goal of this paper is to overcome this problem
by providing a novel information-maximization clustering method.
More specifically, we propose to employ a variant of MI called \emph{squared-loss MI} (SMI),
and develop a new clustering algorithm whose solution can be computed analytically
in a computationally efficient way via kernel eigenvalue decomposition.
Furthermore, for kernel parameter optimization,
we propose to use a non-parametric SMI estimator
called \emph{least-squares MI} \citep[LSMI;][]{BMCBio:Suzuki+etal:2009},
which was proved to achieve the optimal convergence rate
with analytic-form solutions.
Through experiments on various real-world datasets
such as images, natural languages, accelerometric sensors, and speech,
we demonstrate the usefulness of the proposed clustering method.

The rest of this paper is structured as follows.
In Section~\ref{sec:SMIclustering},
we describe our proposed information-maximization clustering method
based on SMI.
Then the proposed method is compared with
existing clustering methods 
qualitatively in Section~\ref{sec:existing}
and quantitatively in Section~\ref{sec:experiments}.
Finally, this paper is concluded in Section~\ref{sec:conclusion}.

\section{Information-Maximization Clustering with Squared-Loss Mutual Information}
\label{sec:SMIclustering}

In this section, we describe our proposed clustering algorithm.

\subsection{Formulation of Information-Maximization Clustering}
\label{sec:MIC}

Suppose we are given $\inputdim$-dimensional i.i.d.~feature vectors of size $\nsample$,
\begin{align*}
  \{\boldx_i\;|\;\boldx_i\in\mathbbR^\inputdim\}_{i=1}^\nsample,
\end{align*}
which are assumed to be drawn independently from a distribution
with density $\density(\boldx)$.
The goal of clustering is to give cluster assignments,
\begin{align*}
  \{y_i\;|\;y_i\in\{1,\ldots,c\}\}_{i=1}^\nsample,
\end{align*}
to the feature vectors $\{\boldx_i\}_{i=1}^\nsample$,
where $c$ denotes the number of classes.
Throughout this paper,
we assume that $c$ is known.

In order to solve the clustering problem,
we take the \emph{information-maximization} approach
\citep{NIPS2005_569,NIPS2010_0457}.
That is, we regard clustering as an unsupervised classification problem,
and learn the class-posterior probability $\density(y|\boldx)$
so that `information' between feature vector $\boldx$ and class label $y$ is maximized.

The \emph{dependence-maximization} approach 
\citep[][see also Section~\ref{subsec:dependence-maximization}]{ICML:Song+etal:2007b,ICML:Faivishevsky+Goldberger:2010}
is related to, but substantially different from the above
information-maximization approach.
In the dependence-maximization approach,
cluster assignments $\{y_i\}_{i=1}^\nsample$ are directly determined
so that their dependence on feature vectors $\{\boldx_i\}_{i=1}^\nsample$ is maximized.
Thus, the dependence-maximization approach intrinsically involves combinatorial optimization
with respect to $\{y_i\}_{i=1}^\nsample$.
On the other hand, the information-maximization approach
involves continuous optimization with respect to 
the parameter $\boldalpha$ included in a class-posterior model
$\densitymodel(y|\boldx;\boldalpha)$.
This continuous optimization of $\boldalpha$ is substantially easier to solve
than discrete optimization of $\{y_i\}_{i=1}^\nsample$.

Another advantage of the information-maximization approach
is that it naturally allows out-of-sample clustering based on the discriminative
model $\densitymodel(y|\boldx;\boldalpha)$,
i.e., a cluster assignment for a new feature vector can be obtained
based on the learned discriminative model.


\subsection{Squared-Loss Mutual Information}
As an information measure, we adopt \emph{squared-loss mutual information} (SMI).
SMI between feature vector $\boldx$ and class label $y$
is defined by
\begin{align}
  \mathrm{SMI}&:=\frac{1}{2}\int\sum_{y=1}^{\nclass} \density(\boldx)\density(y)
  \left(\frac{\density(\boldx,y)}{\density(\boldx)\density(y)}-1\right)^2
  \mathrm{d}\boldx,
  \label{SMI}
\end{align}
where $\density(\boldx,y)$ denotes the joint density of $\boldx$ and $y$,
and $\density(y)$ is the marginal probability of $y$.
SMI is the \emph{Pearson divergence} \citep{PhMag:Pearson:1900} 
from $\density(\boldx,y)$ to $\density(\boldx)\density(y)$,
while the ordinary MI \citep{book:Cover+Thomas:2006},
\begin{align}
  \mathrm{MI}:=\int\sum_{y=1}^{\nclass} \density(\boldx,y)
  \log\frac{\density(\boldx,y)}{\density(\boldx)\density(y)}
  \mathrm{d}\boldx,
\label{MI}
\end{align}
is the \emph{Kullback-Leibler divergence}
\citep{Annals-Math-Stat:Kullback+Leibler:1951}
from $\density(\boldx,y)$ to $\density(\boldx)\density(y)$.
The Pearson divergence and the Kullback-Leibler divergence both 
belong to the class of \emph{Ali-Silvey-Csisz\'ar divergences}
\citep[which is also known as \emph{$f$-divergences}, see][]{JRSS-B:Ali+Silvey:1966,SSM-Hungary:Csiszar:1967},
and thus they share similar properties.
For example, SMI is non-negative and takes zero
if and only if $\boldx$ and $y$ are statistically independent,
as the ordinary MI.

In the existing information-maximization clustering methods
\citep[][see also Section~\ref{subsec:MIclustering}]{NIPS2005_569,NIPS2010_0457},
MI is used as the information measure.
On the other hand, in this paper, we adopt SMI
because it allows us to develop
a clustering algorithm whose solution can be computed analytically
in a computationally efficient way via kernel eigenvalue decomposition.

\subsection{Clustering by SMI Maximization}
\label{subsec:SMIC}
Here, we give a computationally-efficient clustering algorithm
based on SMI \eqref{SMI}.

Expanding the squared term in Eq.\eqref{SMI},
we can express SMI as
\begin{align}
  \mathrm{SMI}
  &=
  \frac{1}{2}\int\sum_{y=1}^{\nclass} \density(\boldx)\density(y)
  \left(\frac{\density(\boldx,y)}{\density(\boldx)\density(y)}\right)^2
  \mathrm{d}\boldx\nonumber\\ 
  &\phantom{=}
  -\int\sum_{y=1}^{\nclass} \density(\boldx)\density(y)
  \frac{\density(\boldx,y)}{\density(\boldx)\density(y)}
  \mathrm{d}\boldx+\frac{1}{2}\nonumber\\
  &=
  \frac{1}{2}\int\sum_{y=1}^{\nclass} \density(y|\boldx)\density(\boldx)
  \frac{\density(y|\boldx)}{\density(y)}
  \mathrm{d}\boldx -\frac{1}{2}.
  \label{SMI4}
\end{align}
Suppose that the class-prior probability $\density(y)$ is set to
a user-specified value $\pi_y$ for $y=1,\ldots,\nclass$,
where $\pi_y>0$ and $\sum_{y=1}^{\nclass}\pi_y=1$.
Without loss of generality, we assume that $\{\pi_y\}_{y=1}^{\nclass}$ are sorted
in the ascending order:
\begin{align*}
  \pi_1\le\cdots\le\pi_{\nclass}.
\end{align*}
If $\{\pi_y\}_{y=1}^{\nclass}$ is unknown, we may merely adopt
the uniform class-prior distribution:
\begin{align}
  \density(y)=\frac{1}{\nclass} \mbox{ for } y=1,\ldots,\nclass,
  \label{p(y)-uniform}
\end{align}
which will be non-informative and thus allow us to avoid biasing clustering solutions\footnote{
Such a cluster-balance constraint is often employed
in existing clustering algorithms
\citep[e.g.,][]{IEEE-PAMI:Shi+Malik:2000,NIPS2005_834,AISTATS:Niu+etal:2011}.
}.
Substituting $\pi_y$ into $\density(y)$,
we can express Eq.\eqref{SMI4} as
\begin{align}
  \frac{1}{2}\int\sum_{y=1}^{\nclass} \frac{1}{\pi_y}\density(y|\boldx)\density(\boldx)
  \density(y|\boldx)
  \mathrm{d}\boldx -\frac{1}{2}.
  \label{SMI2}
\end{align}

Let us approximate the class-posterior probability $\density(y|\boldx)$ by
the following kernel model:
\begin{align}
  \densitymodel(y|\boldx;\boldalpha)&:=\sum_{i=1}^{\nsample}\alpha_{y,i} K_{}(\boldx,\boldx_i),
  \label{kernel-model}
\end{align}
where $\boldalpha=(\alpha_{1,1},\ldots,\alpha_{\nclass,\nsample})^\top$ is the parameter vector,
$^\top$ denotes the transpose,
and $K_{}(\boldx,\boldx')$ denotes a kernel function
with a kernel parameter $\kernelparameter$.
In the experiments, we will use a sparse variant of
the \emph{local-scaling kernel} \citep{NIPS17:Zelnik-Manor+Perona:2005}:
\begin{align}
  K_{}(\boldx_i,\boldx_j)
  =
  \begin{cases}
    \displaystyle
    \exp\left(-\frac{\|\boldx_i-\boldx_j\|^2}{2\sigma_i\sigma_j}\right)
    & \mbox{if $\boldx_i\in \calN_\kernelparameter(\boldx_j)$
      or $\boldx_j\in \calN_\kernelparameter(\boldx_i)$},\\[5mm]
    0 & \mbox{otherwise},
  \end{cases}
\label{sparse-local-scaling}
\end{align}
where $\calN_\kernelparameter(\boldx)$ denotes the set of
$\kernelparameter$ nearest neighbors for $\boldx$
($\kernelparameter$ is the kernel parameter),
$\sigma_i$ is a local scaling factor defined as
$\sigma_i=\|\boldx_i-\boldx_i^{(\kernelparameter)}\|$,
and $\boldx_i^{(\kernelparameter)}$ is
the $\kernelparameter$-th nearest neighbor of $\boldx_i$.

Further approximating the expectation with respect to $\density(\boldx)$
included in Eq.\eqref{SMI2}
by the empirical average of samples $\{\boldx_i\}_{i=1}^{\nsample}$,
we arrive at the following SMI approximator:
\begin{align}
  \widehat{\mathrm{SMI}}
  &:=
  \frac{1}{2\nsample}\sum_{y=1}^{\nclass}\frac{1}{\pi_y}\boldalpha_y^\top\boldK^2\boldalpha_y
  -\frac{1}{2},
  \label{SMIhat}
\end{align}
where $\boldalpha_y:=(\alpha_{y,1},\ldots,\alpha_{y,\nsample})^\top$
and
$K_{i,j}:=K_{}(\boldx_i,\boldx_j)$.

For each cluster $y$, we maximize $\boldalpha_y^\top\boldK^2\boldalpha_y$
under\footnote{Note that this unit-norm constraint is not essential
since the obtained solution is renormalized later.}
 $\|\boldalpha_y\|=1$.
Since this is the \emph{Rayleigh quotient},
the maximizer is given by the normalized principal eigenvector of $\boldK$
\citep{Book:Horn+Johnson:1985}.
To avoid all the solutions $\{\boldalpha_y\}_{y=1}^{\nclass}$ to be reduced
to the same principal eigenvector,
we impose their mutual orthogonality: $\boldalpha_y^\top\boldalpha_{y'}=0$ for $y\neq y'$.
Then the solutions are given by
the normalized eigenvectors $\boldphi_1,\ldots,\boldphi_\nclass$ associated with the eigenvalues
$\lambda_1\ge\cdots\ge\lambda_\nsample\ge0$ of $\boldK$.
Since the sign of $\boldphi_y$ is arbitrary, we set the sign as
\begin{align*}
  \widetilde{\boldphi}_y=\boldphi_y\times\mathrm{sign}(\boldphi_y^\top\boldone_{\nsample}),
\end{align*}
where $\mathrm{sign}(\cdot)$ denotes the sign of a scalar
and
$\boldone_{\nsample}$ denotes the $\nsample$-dimensional vector with all ones.

On the other hand,
since
\begin{align*}
  \density(y)=\int\density(y|\boldx)\density(\boldx)\mathrm{d}\boldx
  \approx\frac{1}{\nsample}\sum_{i=1}^\nsample \densitymodel(y|\boldx_i;\boldalpha)
  =\boldalpha_y^\top\boldK\boldone_{\nsample},
\end{align*}
and the class-prior probability $\density(y)$ was set to
$\pi_y$ for $y=1,\ldots,\nclass$,
we have the following normalization condition:
\begin{align*}
  \boldalpha_y^\top\boldK\boldone_{\nsample}=\pi_y.
\end{align*}
Furthermore, probability estimates should be non-negative,
which can be achieved by rounding up negative outputs to zero.

Taking these normalization and non-negativity issues into account, 
cluster assignment $y_i$ for $\boldx_i$ is determined as
the maximizer of the approximation of $p(y|\boldx_i)$:
\begin{align*}
  y_i&=\argmax_y 
  \frac{[\max(\boldzero_{\nsample},\boldK\widetilde{\boldphi}_y)]_i}
  {\pi_y^{-1}\max(\boldzero_{\nsample},\boldK\widetilde{\boldphi}_y)^\top\boldone_{\nsample}}
  =\argmax_y 
  \frac{\pi_y[\max(\boldzero_{\nsample},\widetilde{\boldphi}_y)]_i}
  {\max(\boldzero_{\nsample},\widetilde{\boldphi}_y)^\top\boldone_{\nsample}},
\end{align*}
where the max operation for vectors is applied in the element-wise manner
and $[\cdot]_i$ denotes the $i$-th element of a vector.
Note that we used $\boldK\widetilde{\boldphi}_y=\lambda_y\widetilde{\boldphi}_y$
in the above derivation.
For out-of-sample prediction, cluster assignment $y'$
for new sample $\boldx'$ may be obtained as
\begin{align*}
  y'&:=
  \argmax_y 
  \frac{\pi_y\max\left(0,\sum_{i=1}^\nsample K(\boldx',\boldx_i)[\widetilde{\boldphi}_y]_i\right)}
  {\lambda_y\max(\boldzero_{\nsample},\widetilde{\boldphi}_y)^\top\boldone_{\nsample}}.
\end{align*}


We call the above method \emph{SMI-based clustering} (SMIC).

\subsection{Kernel Parameter Choice by SMI Maximization}
\label{subsec:LSMI}
The solution of SMIC depends on the choice of
the kernel parameter $\kernelparameter$
included in the kernel function $K_{}(\boldx,\boldx')$.
Since SMIC was developed in the framework of SMI maximization,
it would be natural to determine the kernel parameter $\kernelparameter$
so as to maximize SMI.
A direct approach is to use the SMI estimator $\widehat{\mathrm{SMI}}$ \eqref{SMIhat}
also for kernel parameter choice.
However, this direct approach is not favorable
because $\widehat{\mathrm{SMI}}$ is an unsupervised SMI estimator
(i.e., SMI is estimated only from unlabeled samples $\{\boldx_i\}_{i=1}^{\nsample}$).
On the other hand, in the model selection stage, we have already obtained 
labeled samples $\{(\boldx_i,y_i)\}_{i=1}^{\nsample}$,
and thus supervised estimation of SMI is possible.
For supervised SMI estimation, a non-parametric SMI estimator
called \emph{least-squares mutual information} \citep[LSMI;][]{BMCBio:Suzuki+etal:2009}
was shown to achieve the optimal convergence rate.
For this reason, we propose to use LSMI for model selection,
instead of $\widehat{\mathrm{SMI}}$ \eqref{SMIhat}.

LSMI is an estimator of SMI based on paired samples $\{(\boldx_i,y_i)\}_{i=1}^{\nsample}$.
The key idea of LSMI is to learn the following \emph{density-ratio function},
\begin{align}
\ratio(\boldx,y) := \frac{\density(\boldx,y)}{\density(\boldx)\density(y)},
\label{density-ratio}
\end{align}
without going through density estimation of
$\density(\boldx,y)$, $\density(\boldx)$, and $\density(y)$.
More specifically,
let us employ the following density-ratio model:
\begin{align}
\ratiomodel(\boldx,y;\boldtheta) &:=\sum_{\ell:y_\ell=y}\theta_{\ell}
L_{}(\boldx,\boldx_\ell),
\label{ratio-model}
\end{align}
where $\boldtheta=(\theta_1,\ldots,\theta_{\nsample})^\top$
and
$L_{}(\boldx,\boldx')$ is a kernel function
with a kernel parameter $\kernelparameterr$.
In the experiments, we will use the Gaussian kernel:
\begin{align}
  L_{}(\boldx,\boldx')=
  \exp\left(-\frac{\|\boldx-\boldx'\|^2}{2\kernelparameterr^2}\right),
  \label{Gaussian-kernel}
\end{align}
where the Gaussian width $\kernelparameterr$ is the kernel parameter.

The parameter $\boldtheta$ in the above density-ratio model is learned so that
the following squared error is minimized:
\begin{align}
  \min_\boldtheta
  \frac{1}{2}\int\sum_{y=1}^{\nclass}\Big(\ratiomodel(\boldx,y;\boldtheta)-\ratio(\boldx,y)\Big)^2
  \density(\boldx)\density(y)\mathrm{d}\boldx.
  \label{LSMI-squared-error}
\end{align}
Let $\boldtheta_y$ be the parameter vector
corresponding to the kernel bases 
$\{L_{}(\boldx,\boldx_\ell)\}_{\ell:y_\ell=y}$,
i.e., $\boldtheta_y$ is the sub-vector of $\boldtheta=(\theta_1,\ldots,\theta_\nsample)^\top$
consisting of indices $\{\ell\;|\;y_\ell=y\}$.
Let $\nsample_y$ be the length of $\boldtheta_y$,
i.e., the number of samples in cluster $y$.
Then an empirical and regularized version
of the optimization problem \eqref{LSMI-squared-error} is given
for each $y$ as follows:
\begin{align}
  \min_{\boldtheta_y}
  \left[
    \frac{1}{2}\boldtheta_y^\top\boldHh^{(y)}\boldtheta_y
  -\boldtheta_y^\top\boldhh^{(y)}+\frac{\regularizationparameterr}{2}\boldtheta_y^\top\boldtheta_y
  \right],
\label{LSMI-objective}
\end{align}
where $\regularizationparameterr$ ($\ge0$) is the regularization parameter.
$\boldHh^{(y)}$ is the $\nsample_y\times\nsample_y$ matrix
and $\boldhh^{(y)}$ is the $\nsample_y$-dimensional vector defined as
\begin{align*}
\Hh_{\ell,\ell'}^{(y)}&:=\frac{\nsample_y}{\nsample^2}
\sum_{i=1}^\nsample L_{}(\boldx_i,\boldx_{\ell}^{(y)})
L_{}(\boldx_i,\boldx_{\ell'}^{(y)}),\\
\hh_{\ell}^{(y)}&:=\frac{1}{\nsample}
\sum_{i:y_i=y}L_{}(\boldx_i,\boldx_{\ell}^{(y)}),
\end{align*}
where $\boldx^{(y)}_\ell$ is the $\ell$-th sample in class $y$
(which corresponds to $\thetah^{(y)}_{\ell}$).

A notable advantage of LSMI is that 
the solution $\boldthetah^{(y)}$ can be computed analytically as
\begin{align*}
  \boldthetah^{(y)}=(\boldHh^{(y)}+\regularizationparameterr\boldI)^{-1}\boldhh^{(y)}.
\end{align*}
Then a density-ratio estimator is obtained analytically as follows\footnote{
Note that, in the original LSMI paper \citep{BMCBio:Suzuki+etal:2009},
the entire parameter $\boldtheta=(\theta_1,\ldots,\theta_\nsample)^\top$ 
for all classes was optimized at once.
On the other hand, we found that,
when the density-ratio model $\ratiomodel(\boldx,y;\boldtheta)$ defined by Eq.\eqref{ratio-model}
is used for SMI approximation,
exactly the same solution as the original LSMI paper
can be computed more efficiently by class-wise optimization.
Indeed, in our preliminary experiments, we confirmed that
our class-wise optimization significantly reduces
the computation time compared with the original all-class optimization,
with the same solution.
Note that the original LSMI is applicable to 
more general setups such as regression, multi-label classification,
and structured-output prediction.
Thus, our speedup was brought by focusing on classification scenarios
where Kronecker's delta function is used as the kernel for class labels
in the density-ratio model \eqref{ratio-model}.
}:
\begin{align*}
\ratioh(\boldx,y)&=\sum_{\ell=1}^{\nsample_y}\thetah^{(y)}_{\ell}
L_{}(\boldx,\boldx^{(y)}_\ell).
\end{align*}

The accuracy of the above least-squares density-ratio estimator depends on 
the choice of the kernel parameter $\kernelparameterr$ included in $L_{}(\boldx,\boldx')$
and the regularization parameter $\regularizationparameterr$ in Eq.\eqref{LSMI-objective}.
\citet{BMCBio:Suzuki+etal:2009}
showed that these tuning parameter values 
can be systematically optimized based on cross-validation as follows:
First, the samples $\calZ=\{(\boldx_i,y_i)\}_{i=1}^{\nsample}$
are divided into $M$ disjoint subsets $\{\calZ_m\}_{m=1}^M$
of approximately the same size (we use $M=5$ in the experiments).
Then a density-ratio estimator $\ratioh_m(\boldx,y)$
is obtained using $\calZ\backslash\calZ_m$
(i.e., all samples without $\calZ_m$),
and its out-of-sample error (which corresponds to Eq.\eqref{LSMI-squared-error}
without irrelevant constant)
for the hold-out samples $\calZ_m$ is computed as
\begin{align*}
  \mathrm{CV}_m:=\frac{1}{2|\calZ_m|^2}\sum_{\boldx,y\in\calZ_m}\ratioh_m(\boldx,y)^2
  -\frac{1}{|\calZ_m|}\sum_{(\boldx,y)\in\calZ_m}\ratioh_m(\boldx,y).
\end{align*}
This procedure is repeated for $m=1,\ldots,M$,
and the average of the above hold-out error
over all $m$ is computed as
\begin{align*}
  \mathrm{CV}:=\frac{1}{M}\sum_{m=1}^M  \mathrm{CV}_m.
\end{align*}
Finally, the kernel parameter $\kernelparameterr$
and the regularization parameter $\regularizationparameterr$
that minimize the average hold-out error $\mathrm{CV}$
are chosen as the most suitable ones.

Finally, based on an expression of SMI \eqref{SMI},
\begin{align*}
  \mathrm{SMI}&=
  -\frac{1}{2}\int\sum_{y=1}^{\nclass}\ratio(\boldx,y)^2\density(\boldx)\density(y)\mathrm{d}\boldx
+ \int\sum_{y=1}^{\nclass}\ratio(\boldx,y)\density(\boldx,y)\mathrm{d}\boldx
-\frac{1}{2},
\end{align*}
an SMI estimator called LSMI is given as follows:
\begin{align}
  \mathrm{LSMI}:=
  -\frac{1}{2\nsample^2}\sum_{i,j=1}^{\nsample}\ratioh(\boldx_i,y_j)^2
    +\frac{1}{\nsample}\sum_{i=1}^{\nsample}\ratioh(\boldx_i,y_i)-\frac{1}{2},
\label{LSMI}
\end{align}
where $\ratioh(\boldx,y)$ is a density-ratio estimator obtained above.
Since $\ratioh(\boldx,y)$ can be computed analytically,
LSMI can also be computed analytically.

We use LSMI for model selection of SMIC.
More specifically, we compute LSMI as a function of the kernel parameter $\kernelparameter$
of $K_{}(\boldx,\boldx')$ included in the cluster-posterior model \eqref{kernel-model},
and choose the one that maximizes LSMI.
A pseudo code of the entire SMI-maximization clustering procedure
is summarized in 
Figures~\ref{fig:SMIC-LSMI}--\ref{fig:LSMI}.
Its MATLAB implementation is available from
\begin{center}
  `\url{http://sugiyama-www.cs.titech.ac.jp/~sugi/software/SMIC}'.
\end{center}

\begin{figure}[t]
  \centering
   \framebox{
    \begin{minipage}{0.6\linewidth}
      \begin{tabbing}
        XX\=XX\=\kill
        \textbf{Input:} Feature vectors $\calX=\{\boldx_i\}_{i=1}^\nsample$
        and the number $\nclass$ of clusters\\
        \textbf{Output:} Cluster assignments $\calY=\{y_i\}_{i=1}^\nsample$\\[2mm]
        \textbf{For} each kernel parameter candidate $\kernelparameter\in\Kernelparameter$\\
        \>$\calY^{(\kernelparameter)}\longleftarrow\mathrm{SMIC}(\calX,\kernelparameter,\nclass)$;\\
        \>$\mathrm{LSMI}(\kernelparameter)\longleftarrow
        \mathrm{LSMI}(\calX,\calY^{(\kernelparameter)})$;\\
        \textbf{end}\\
        $\displaystyle\widehat{\kernelparameter}\longleftarrow\argmax_{\kernelparameter\in\Kernelparameter}
        \mathrm{LSMI}(\kernelparameter)$;\\
        $\calY\longleftarrow \calY^{(\widehat{\kernelparameter})}$;
      \end{tabbing}
    \end{minipage}
    }
\caption{Pseudo code of information-maximization clustering based on SMIC and LSMI.
The kernel parameter $\kernelparameter$ refers to the tuning parameter
included in the kernel function $K_{}(\boldx,\boldx')$
in the cluster-posterior model \eqref{kernel-model}.
Pseudo codes of SMIC and LSMI are described in
Figure~\ref{fig:SMIC} and Figure~\ref{fig:LSMI}, respectively.
}
\label{fig:SMIC-LSMI}
 \end{figure}

\begin{figure}[p]
  \centering
   \framebox{
    \begin{minipage}{0.6\linewidth}
      \begin{tabbing}
        XX\=XX\=\kill
        \textbf{Input:} Feature vectors $\calX=\{\boldx_i\}_{i=1}^\nsample$,
        kernel parameter $\kernelparameter$,\\
        \phantom{\textbf{Input:}}
        and the number $\nclass$ of clusters\\
        \textbf{Output:} Cluster assignments $\calY=\{y_i\}_{i=1}^\nsample$\\[2mm]
        $\boldK\longleftarrow$ Kernel matrix for 
        samples $\calX$ and kernel parameter $\kernelparameter$;\\
        $\boldphi_y\longleftarrow$
        $y$-th principal eigenvectors of $\boldK$ for $y=1,\ldots,\nclass$;\\
        $\widetilde{\boldphi}_y\longleftarrow\boldphi_y\times
        \mathrm{sign}(\boldphi_y^\top\boldone_{\nsample})$ for $y=1,\ldots,\nclass$;\\
        $\displaystyle y_i\longleftarrow\argmax_{y\in\{1,\ldots,\nclass\}} 
        \frac{[\max(\boldzero_{\nsample},\widetilde{\boldphi}_y)]_i}
        {\max(\boldzero_{\nsample},\widetilde{\boldphi}_y)^\top\boldone_{\nsample}}$
        for $i=1,\ldots,\nsample$;\\
        $\calY\longleftarrow\{y_i\}_{i=1}^\nsample$;
      \end{tabbing}
    \end{minipage}
    }
\caption{Pseudo code of SMIC (with the uniform class-prior distribution).
The kernel parameter $\kernelparameter$ refers to the tuning parameter
included in the kernel function $K_{}(\boldx,\boldx')$
in the cluster-posterior model \eqref{kernel-model}.
If the class-prior probability $\density(y)$ is set to
a user-specified value $\pi_y$ for $y=1,\ldots,\nclass$,
$y_i$ is determined as
$\argmax_y  \frac{\pi_y[\max(\boldzero_{\nsample},\widetilde{\boldphi}_y)]_i}
{\max(\boldzero_{\nsample},\widetilde{\boldphi}_y)^\top\boldone_{\nsample}}$.
}
\label{fig:SMIC}
\vspace*{10mm}
   \framebox{
    \begin{minipage}{0.6\linewidth}
      \begin{tabbing}
        XX\=XX\=XX\=XX\=\kill
        \textbf{Input:} Feature vectors $\calX=\{\boldx_i\}_{i=1}^\nsample$
        and cluster assignments $\calY=\{y_i\}_{i=1}^\nsample$\\
        \textbf{Output:} SMI estimate $\mathrm{LSMI}$\\[2mm]
        $\calZ\longleftarrow\{(\boldx_i,y_i)\}_{i=1}^{\nsample}$;\\
        $\{\calZ_m\}_{m=1}^M\longleftarrow$ $M$ disjoint subsets of $\calZ$;\\
        \textbf{For} each kernel parameter candidate $\kernelparameterr\in\Kernelparameterr$\\
        \>\textbf{For} each regularization parameter candidate
        $\regularizationparameterr\in\Regularizationparameterr$\\
        \>\>\textbf{For} each fold $m=1,\ldots,M$\\
        \>\>\>$\ratioh_{\kernelparameterr,\regularizationparameterr,m}(\boldx,y)\longleftarrow$ Density ratio estimator
        for $(\kernelparameterr,\regularizationparameterr)$ using $\calZ\backslash\calZ_m$;\\
        \>\>\>$\mathrm{CV}_m(\kernelparameterr,\regularizationparameterr)\longleftarrow$
        Hold-out error of $\ratioh_{\kernelparameterr,\regularizationparameterr,m}(\boldx,y)$
        for $\calZ_m$;\\
        \>\>\textbf{end}\\
        \>\>$\displaystyle\mathrm{CV}(\kernelparameterr,\regularizationparameterr)\longleftarrow
        \frac{1}{M}\sum_{m=1}^M\mathrm{CV}_m(\kernelparameterr,\regularizationparameterr)$;\\
        \>\textbf{end}\\
        \textbf{end}\\
        $\displaystyle
        (\widehat{\kernelparameterr},\widehat{\regularizationparameterr})\longleftarrow
        \argmin_{\kernelparameterr\in\Kernelparameterr,\regularizationparameterr\in\Regularizationparameterr}
        \mathrm{CV}(\kernelparameterr,\regularizationparameterr)$;\\
        $\ratioh(\boldx,y)\longleftarrow$ Density ratio estimator
        for $(\widehat{\kernelparameterr}, \widehat{\regularizationparameterr})$ using $\calZ$;\\
        $\displaystyle \mathrm{LSMI}\longleftarrow
  -\frac{1}{2\nsample^2}\sum_{i,j=1}^{\nsample}\ratioh(\boldx_i,y_j)^2
    +\frac{1}{\nsample}\sum_{i=1}^{\nsample}\ratioh(\boldx_i,y_i)-\frac{1}{2},
$;
      \end{tabbing}
    \end{minipage}
  }
\caption{Pseudo code of LSMI.
The kernel parameter $\kernelparameterr$ refers to the tuning parameter
included in the kernel function $L_{}(\boldx,\boldx')$
in the density-ratio model \eqref{ratio-model}.
}
\label{fig:LSMI}
\end{figure}

\section{Existing Clustering Methods}
\label{sec:existing}

In this section, we review existing clustering methods
and qualitatively discuss the relation to the proposed approach.

\subsection{K-Means Clustering}
\label{subsec:k-means}
\emph{K-means clustering} \citep{BerkeleySymp:MacQueen:1967}
would be one of the most popular clustering algorithms.
It tries to minimize the following distortion measure
with respect to the cluster assignments $\{y_i\}_{i=1}^\nsample$:
\begin{align}
  \sum_{y=1}^\nclass \sum_{i:y_i=y}\|\boldx_i-\boldmu_y\|^2,
  \label{k-means}
\end{align}
where
$\boldmu_y:=\frac{1}{\nsample_y}\sum_{i:y_i=y}\boldx_i$
is the centroid of cluster $y$ and
$\nsample_y$ is the number of samples in cluster $y$.

The original k-means algorithm is capable of only producing linearly separated clusters
\citep{book:Duda+etal:2001}.
However, since samples are used only in terms of their inner products,
its non-linear variant can be
immediately obtained by performing k-means in 
a feature space induced by a reproducing kernel function
\citep{IEEE-TNN:Girolami:2002}.

As the optimization problem of (kernel) k-means is NP-hard
\citep{MLJ:Aloise+etal:2009},
a greedy optimization algorithm is usually used
for finding a local optimal solution in practice.
It was shown that
the solution to a continuously-relaxed variant of the kernel k-means problem
is given by the principal components of the kernel matrix
\citep{nips02-AA41,ICML:Ding+He:2004}.
Thus, post-discretization of the relaxed solution may give
a good approximation to the original problem,
which is computationally efficient.
This idea is similar to the proposed SMIC method described in Section~\ref{subsec:SMIC}.
However, an essential difference is that
SMIC handles the continuous solution directly
as a parameter estimate of the class-posterior model.

The performance of kernel k-means depends heavily on
the choice of kernel functions,
and there is no systematic way to determine the kernel function.
This is a critical weakness of kernel k-means in practice.
On the other hand, our proposed approach offers a natural model selection strategy,
which is a significant advantage over kernel k-means.

\subsection{Spectral Clustering}
\label{subsec:spectral-clustering}
The basic idea of \emph{spectral clustering} \citep{IEEE-PAMI:Shi+Malik:2000,nips02-AA35}
is to first unfold non-linear data manifolds 
by a spectral embedding method,
and then perform k-means in the embedded space.
More specifically, given sample-sample similarity $W_{i,j}\ge0$
(large $W_{i,j}$ means that $\boldx_i$ and $\boldx_j$ are similar),
the minimizer of the following criterion
with respect to $\{\boldxi_i\}_{i=1}^\nsample$
is obtained under some normalization constraint:
\begin{align*}
  \sum_{i,j}^{\nsample} W_{i,j}\left\|\frac{1}{\sqrt{D_{i,i}}}\boldxi_i
    -\frac{1}{\sqrt{D_{j,j}}}\boldxi_j\right\|^2,
\end{align*}
where $\boldD$ is the diagonal matrix with
$i$-th diagonal element given by $D_{i,i}:=\sum_{j=1}^{\nsample}W_{i,j}$.
Consequently, the embedded samples are given by the principal eigenvectors of
$\boldD^{-\frac{1}{2}}\boldW\boldD^{-\frac{1}{2}}$, followed by normalization.
Note that spectral clustering was shown to be
equivalent to a weighted variant of kernel k-means
with some specific kernel \citep{KDD:Dhillon+etal:2004}.

The performance of spectral clustering depends heavily on
the choice of sample-sample similarity $W_{i,j}$.
\citet{NIPS17:Zelnik-Manor+Perona:2005} proposed a
useful unsupervised heuristic to determine the similarity
in a data-dependent manner, called \emph{local scaling}:
\begin{align*}
  W_{i,j}  = \exp\left(-\frac{\|\boldx_i-\boldx_j\|^2}{2\sigma_i\sigma_j}\right),
\end{align*}
where $\sigma_i$ is a local scaling factor defined as
\begin{align*}
  \sigma_i=\|\boldx_i-\boldx_i^{(\kernelparameter)}\|,
\end{align*}
and $\boldx_i^{(\kernelparameter)}$ is the $\kernelparameter$-th nearest neighbor of $\boldx_i$.
$\kernelparameter$ is the tuning parameter in the local scaling similarity,
and $\kernelparameter=7$ was shown to be useful
\citep{NIPS17:Zelnik-Manor+Perona:2005,JMLR:Sugiyama:2007}.
However, this magic number `7' does not seem to work always well in general.

If $\boldD^{-\frac{1}{2}}\boldW\boldD^{-\frac{1}{2}}$ is regarded as a kernel matrix,
spectral clustering will be
similar to the proposed SMIC method described in Section~\ref{subsec:SMIC}.
However, SMIC does not require the post k-means processing
since the principal components have clear interpretation
as parameter estimates of the class-posterior model \eqref{kernel-model}.
Furthermore, our proposed approach
provides a systematic model selection strategy,
which is a notable advantage over spectral clustering.

\subsection{Blurring Mean-Shift Clustering}
\emph{Blurring mean-shift} \citep{IEEE-IT:Fukunaga+Hostetler:1975}
is a non-parametric clustering method based on
the \emph{modes} of the data-generating probability density.

In the blurring mean-shift algorithm,
a kernel density estimator \citep{book:Silverman:1986}
is used for modeling the data-generating probability density:
\begin{align*}
  \widehat{p}(\boldx)=\frac{1}{\nsample}\sum_{i=1}^\nsample 
  K\left({\left\|\boldx-\boldx_i\right\|^2}/{\sigma^2}\right),
\end{align*}
where $K(\xi)$ is a kernel function
such as a Gaussian kernel $K(\xi)=e^{-\xi/2}$.
Taking the derivative of $\widehat{p}(\boldx)$ with respect to $\boldx$
and equating the derivative at $\boldx=\boldx_i$ to zero,
we obtain the following updating formula for sample $\boldx_i$ ($i=1,\ldots,\nsample$):
\begin{align*}
   \boldx_i\longleftarrow
   \frac{\sum_{j=1}^\nsample W_{i,j}\boldx_j}
   {\sum_{j'=1}^\nsample W_{i,j'}},
\end{align*}
where 
$W_{i,j}:=K'\left({\left\|\boldx_i-\boldx_j\right\|^2}/{\sigma^2}\right)$
and $K'(\xi)$ is the derivative of $K(\xi)$.
Each mode of the density is regarded as a representative of a cluster,
and each data point is assigned to the cluster which it converges to.

\citet{IEEE-PAMI:Carreira-Perpinan:2007} showed 
that the blurring mean-shift algorithm
can be interpreted as an \emph{expectation-maximization algorithm} \citep{JRSS-B:Dempster+etal:1977},
where $W_{i,j}/(\sum_{j'=1}^\nsample W_{i,j'})$ is regarded as the posterior probability
of the $i$-th sample belonging to the $j$-th cluster.
Furthermore, the above update rule can be expressed in a matrix form as
$\boldX\longleftarrow\boldX\boldP$,
where
$\boldX=(\boldx_1,\ldots,\boldx_\nsample)$ is a sample matrix
and
$\boldP:=\boldW\boldD^{-1}$
is a \emph{stochastic matrix}
of the random walk in a graph with adjacency $\boldW$ \citep{book:Chung:1997}.
$\boldD$ is defined as
$D_{i,i}:=\sum_{j=1}^{\nsample}W_{i,j}$
and $D_{i,j}=0$ for $i\neq j$.
If $\boldP$ is independent of $\boldX$,
the above iterative algorithm corresponds to the \emph{power method}
\citep{book:Golub+vanLoan:1996} for finding the leading left eigenvector of $\boldP$.
Then, this algorithm is highly related to the spectral clustering
which computes the principal eigenvectors of
$\boldD^{-\frac{1}{2}}\boldW\boldD^{-\frac{1}{2}}$ (see Section~\ref{subsec:spectral-clustering}).
Although $\boldP$ depends on $\boldX$ in reality,
\citet{ICML:Carreira-Perpinan:2006} insisted that this analysis is still valid
since $\boldP$ and $\boldX$ quickly reach a quasi-stable state.

An attractive property of blurring mean-shift
is that the number of clusters is automatically determined
as the number of modes in the probability density estimate.
However, this choice depends on the kernel parameter $\sigma$
and there is no systematic way to determine $\sigma$,
which is restrictive compared with the proposed method.
Another critical drawback of the blurring mean-shift algorithm
is that it eventually converges to a single point 
\citep[i.e., a single cluster, see][for details]{IEEE-PAMI:Cheng:1995},
and therefore a sensible stopping criterion is necessary in practice.
Although \citet{ICML:Carreira-Perpinan:2006} gave a useful heuristic for
stopping the iteration,
it is not clear whether this heuristic always works well in practice.

\subsection{Discriminative Clustering}
\label{subsec:discriminative}
The \emph{support vector machine} \citep[SVM;][]{book:Vapnik:1995}
is a supervised discriminative classifier that tries to find
a hyperplane separating positive and negative samples
with the maximum margin.
\citet{NIPS2005_834} extended SVM
to unsupervised classification scenarios (i.e., clustering),
which is called \emph{maximum-margin clustering} (MMC).

MMC inherits the idea of SVM and tries to find 
the cluster assignments $\boldy=(y_1,\ldots,y_\nsample)^\top$ so that the margin
between two clusters is maximized under proper constraints:
\begin{align*}
  \min_{\boldy\in\{+1,-1\}^\nsample}\max_{\boldlambda}~~&
  2\boldlambda^\top\boldone_\nsample
  -\inner{\boldK\circ\boldlambda\boldlambda^\top}{\boldy\boldy^\top}\\
  \mbox{subject to}~~&
  -\varepsilon\le\boldone_\nsample^\top\boldy\le\varepsilon
  \mbox{ and }
  \boldzero_\nsample\le\boldlambda\le C\boldone_\nsample,
\end{align*}
where $\circ$ denotes the \emph{Hadamard product} (also known as the entry-wise product),
and $\varepsilon$ and $C$ are tuning parameters.
The constraint $-\varepsilon\le\boldone_\nsample^\top\boldy\le\varepsilon$ 
corresponds to balancing the cluster size.

Since the above optimization problem is combinatorial with respect to $\boldy$
and thus hard to solve directly,
it is relaxed to a semi-definite program
by replacing $\boldy\boldy^\top$
(which is a zero-one matrix with rank one)
with a real positive semi-definite matrix
\citep{NIPS2005_834}.
Since then, several approaches have been developed
for further improving the computational efficiency of MMC
\citep{NIPS2006_273,SDM:Zhao+etal:2008,IEEE-TNN:Zhang+etal:2009,AISTATS:Li+etal:2009,IEEE-TNN:Wang+etal:2010}.

The performance of MMC depends heavily on the
choice of the tuning parameters $\varepsilon$ and $C$,
but there is no systematic method to tune these parameters.
The fact that our proposed approach is equipped with a model selection strategy
would practically be a strong advantage over MMC.

Following a similar line to MMC,
a \emph{discriminative and flexible framework for clustering}
\citep[DIFFRAC;][]{NIPS2007_870} was proposed.
DIFFRAC tries to solve a regularized least-squares problem
with respect to a linear predictor and class labels.
Thanks to the simple least-squares formulation,
the parameters in the linear predictor can be optimized analytically,
and thus the optimization problem is much simplified.
A kernelized version of the DIFFRAC optimization problem is given by
\begin{align*}
  \min_{\boldy\in\{+1,-1\}^\nsample}&
  \tr{\boldPi\boldPi^\top\kappa\boldGamma
    (\boldGamma\boldK\boldGamma+\nsample\kappa\boldI_\nsample)^{-1}\boldGamma},
\end{align*}
where $\boldPi$ is the $\nsample\times\nclass$
cluster indicator matrix, which takes $1$
only at one of the elements in each row 
(this corresponds to the index of the cluster to which the sample belongs)
and others are all zeros.
$\kappa$ ($\ge0$) is the regularization parameter,
and
$\boldGamma:=\boldI_\nsample-\frac{1}{\nsample}\boldone_\nsample\boldone_\nsample^\top$
is a centering matrix.
In practice, the above optimization problem is
relaxed to a semi-definite program
by replacing $\boldPi\boldPi^\top$
with a real positive semi-definite matrix.
However, DIFFRAC is still computationally expensive and
it suffers from lack of objective model selection strategies.

\subsection{Generative Clustering}
\label{subsec:generative}
In the \emph{generative clustering} framework \citep{book:Duda+etal:2001},
class labels are determined by
\begin{align*}
  \widehat{y}=\argmax_{y}\density(y|\boldx)
  =\argmax_{y}\density(\boldx,y),
\end{align*}
where $\density(y|\boldx)$ is the class-posterior probability and
$\density(\boldx,y)$ is the data-generating probability.
Typically, $\density(\boldx,y)$ is modeled as
\begin{align*}
  \densitymodel(\boldx,y;\boldbeta,\boldpi)=
  \densitymodel(\boldx|y;\boldbeta)\densitymodel(y;\boldpi),
\end{align*}
where $\boldbeta$ and $\boldpi$ are parameters.
Canonical model choice is the Gaussian distribution
for $\densitymodel(\boldx|y;\boldbeta)$
and the multinomial distribution for $\densitymodel(y;\boldpi)$.

However, since class labels $\{y_i\}_{i=1}^{\nsample}$ are unknown,
one may not directly learn $\boldbeta$ and $\boldpi$
in the joint-probability model $\densitymodel(\boldx,y;\boldbeta,\boldpi)$.
An approach to coping with this problem is to consider a \emph{marginal} model,
\begin{align*}
  \densitymodel(\boldx;\boldbeta,\boldpi)=
  \sum_{y=1}^{\nclass}\densitymodel(\boldx|y;\boldbeta)\densitymodel(y;\boldpi),
\end{align*}
and learns the parameters $\boldbeta$ and $\boldpi$ 
by maximum likelihood estimation \citep{book:Duda+etal:2001}:
\begin{align*}
  \max_{\boldbeta,\boldpi}
  \prod_{i=1}^{\nsample}\densitymodel(\boldx_i;\boldbeta,\boldpi).
\end{align*}
Since the likelihood function of the above mixture model is non-convex,
a \emph{gradient method} \citep{IEEE:Amari:1967} may be used for
finding a local maximizer in practice.
For determining the number of clusters (mixtures)
and the mixing-element model $\densitymodel(\boldx|y;\boldbeta)$,
\emph{likelihood cross-validation} \citep{book:Haerdle+etal:2004}
may be used.

Another approach to coping with the unavailability of class labels
is to regard $\{y_i\}_{i=1}^{\nsample}$ as \emph{latent variables},
and apply the \emph{expectation-maximization (EM) algorithm} \citep{JRSS-B:Dempster+etal:1977}
for finding a local maximizer of the joint likelihood:
\begin{align*}
  \max_{\boldbeta,\boldpi}
  \prod_{i=1}^{\nsample}\densitymodel(\boldx_i,y_i;\boldbeta,\boldpi).
\end{align*}
A more flexible variant of the EM algorithm called
the \emph{split-and-merge EM algorithm} \citep{nc:Ueda+Nakano+Ghahramani:2000}
is also available,
which dynamically controls the number of clusters during the EM iteration.

Instead of point-estimating the parameters $\boldbeta$ and $\boldpi$,
one can also consider their distributions in the \emph{Bayesian} framework
\citep{book:Bishop:2006}.
Let us introduce prior distributions $\densitymodel(\boldbeta)$ and $\densitymodel(\boldpi)$
for the parameters $\boldbeta$ and $\boldpi$.
Then the posterior distribution of the parameters is expressed as
\begin{align*}
  \densitymodel(\boldbeta,\boldpi|\calX)
  \propto
  \densitymodel(\calX|\boldbeta,\boldpi)\densitymodel(\boldbeta)\densitymodel(\boldpi),
\end{align*}
where $\calX=\{\boldx_i\}_{i=1}^\nsample$.
Based on the \emph{Bayesian predictive distribution},
\begin{align*}
  \densityh(y|\boldx,\calX)\propto
\iint
  \densitymodel(\boldx,y|\boldbeta,\boldpi)
  \densitymodel(\boldbeta,\boldpi|\calX)
  \mathrm{d}\boldbeta
  \mathrm{d}\boldpi,
\end{align*}
class labels are determined as
\begin{align*}
  \max_{y}\densityh(y|\boldx,\calX).
\end{align*}

Because the integration included in the Bayesian predictive distribution
is computationally expensive,
\emph{conjugate priors} are often adopted in practice.
For example,
for the Gaussian-cluster model $\densitymodel(\boldx|y;\boldbeta)$,
the Gaussian prior for the mean parameter
and the Wishart prior is assumed for the precision parameter (i.e., the inverse covariance)
are assumed;
the Dirichlet prior is assumed
for the multinomial model $\densitymodel(y;\boldpi)$.
Otherwise, the posterior distribution is approximated by
the \emph{Laplace approximation} \citep{book:MacKay:2003},
the \emph{Markov chain Monte Carlo sampling} \citep{mach:Andrieu+Freitas+Doucet:2003},
or
the \emph{variational approximation} \citep{Att00,GhaBea00}.
The number of clusters can be determined based on
the maximization of the \emph{marginal likelihood}:
\begin{align}
  \densitymodel(\calX)=
\argmax_{y}\iint
  \densitymodel(\calX|\boldbeta,\boldpi)
  \densitymodel(\boldbeta)\densitymodel(\boldpi)
  \mathrm{d}\boldbeta
  \mathrm{d}\boldpi.
  \label{marginal-likelihood}
\end{align}

The generative clustering methods are statistically well-founded.
However, density models for each cluster $\density(\boldx|y)$
need to be specified in advance, which lacks flexibility in practice.
Furthermore, in the Bayesian approach,
the choice of cluster models and prior distributions are often limited
to conjugate pairs in practice.
On the other hand, in the frequentist approach,
only local solutions can be obtained in practice
due to the non-convexity caused by mixture modeling.

\subsection{Posterior-Maximization Clustering}
Another possible clustering approach
based on probabilistic inference is
to directly maximizes the posterior probability
of class labels $\calY=\{y_i\}_{i=1}^\nsample$
\citep{book:Bishop:2006}:
\begin{align*}
  \max_{\calY}\density(\calY|\calX).
\end{align*}
Let us model the cluster-wise data distribution $\density(\calX|\calY)$
by $\densitymodel(\calX|\calY,\boldbeta)$.

An approximate inference method called \emph{iterative conditional modes}
\citep{nc:Kurihara+Welling:2009} alternatively maximizes
the posterior probabilities of $\calY$ and $\boldbeta$
until convergence:
\begin{align*}
  \widehat{\calY}&\longleftarrow \densitymodel(\calY|\calX,\widehat{\boldbeta}),\\
  \widehat{\boldbeta}&\longleftarrow \densitymodel(\boldbeta|\calX,\widehat{\calY}).
\end{align*}
When the Gaussian model with covariance identity
is assumed for $\densitymodel(\calY|\calX,\boldbeta)$,
this algorithm is reduced to the k-means algorithm
(see Section~\ref{subsec:k-means}) under the uniform priors.

Let us consider the class-prior probability $\density(\calY)$ 
and model it by $\densitymodel(\calY|\boldpi)$.
Introducing the prior distributions $\densitymodel(\boldbeta)$ and $\densitymodel(\boldpi)$,
we can approximate the posterior distribution of $\calY$ as
\begin{align*}
  \densitymodel(\calY|\calX)
  \propto
\iint
  \densitymodel(\calX|\calY,\boldbeta)\densitymodel(\boldbeta)
  \densitymodel(\calY|\boldpi)\densitymodel(\boldpi)
  \mathrm{d}\boldbeta
  \mathrm{d}\boldpi.
\end{align*}
Similarly to generative clustering described in Section~\ref{subsec:generative},
conjugate priors such as the Gauss-Wishart prior and the Dirichlet prior
are practically useful in improving the computational efficiency.
The number of clusters can also be similarly determined by maximizing
the marginal likelihood \eqref{marginal-likelihood}.
However, direct optimization of $\calY$ is often computationally
intractable due to $\nclass^{\nsample}$ combinations,
where $\nclass$ is the number of clusters and
$\nsample$ is the number of samples.
For this reason, efficient sampling schemes
such as the Markov chain Monte Carlo are indispensable in this approach.

A \emph{Dirichlet process mixture}
\citep{AS:Ferguson:1973,AS:Antoniak:1974}
is a non-parametric extension of the above approach,
where an infinite number of clusters are implicitly considered
and the number of clusters is automatically determined
based on observed data.
In order to improve the computational efficiency
of this infinite mixture approach,
various approximation schemes
such as Markov chain Monte Carlo sampling \citep{JCGS:Neal:2000}
and variational approximation \citep{BA:Blei+Jordan:2006}
have been introduced.
Furthermore, variants of Dirichlet processes
such as hierarchical Dirichlet processes \citep{JASA:Teh+etal:2007},
nested Dirichlet processes \citep{JASA:Rodriguez+etal:2008},
and dependent Dirichlet processes \citep{NIPS2010_0071}
have been developed recently.

However, even in this non-parametric Bayesian approach, 
density models for each cluster still need to be parametrically specified in advance,
which is often limited to Gaussian models.
This highly limits the flexibility in practice.

\subsection{Dependence-Maximization Clustering}
\label{subsec:dependence-maximization}
The \emph{Hilbert-Schmidt independence criterion}
\citep[HSIC;][]{ALT:Gretton+etal:2005}
is a dependence measure based on a reproducing kernel function $K(\boldx,\boldx')$
\citep{AMS:Aronszajn:1950}.
\citet{ICML:Song+etal:2007b} proposed a \emph{dependence-maximization clustering}
method called \emph{clustering with HSIC} (CLUHSIC),
which tries to determine cluster assignments $\{y_i\}_{i=1}^\nsample$ so that
their dependence on feature vectors $\{\boldx_i\}_{i=1}^\nsample$ is maximized.

More specifically, CLUHSIC tries to find
the cluster indicator matrix $\boldPi$ (see Section~\ref{subsec:discriminative}) that maximizes
\begin{align*}
\tr{\boldK\boldPi\boldA\boldPi^\top},
\end{align*}
where $K_{i,j}:=K(\boldx_i,\boldx_j)$ and
$\boldA$ is a $\nclass\times\nclass$ cluster-cluster similarity matrix.
Note that $\boldPi\boldA\boldPi^\top$ can be regarded 
as the kernel matrix for cluster assignments.
\citet{ICML:Song+etal:2007b} used a greedy algorithm
to optimize the cluster indicator matrix,
which is computationally demanding.
\citet{SDM:Yang+etal:2010} gave spectral and semi-definite
relaxation techniques to improve the computational efficiency of CLUHSIC.

HSIC is a kernel-based independence measure
and the kernel function $K(\boldx,\boldx')$
needs to be determined in advance.
However, there is no systematic model selection strategy for HSIC,
and using the Gaussian kernel with width set to the median distance
between samples is a standard heuristic in practice
\citep{AS:Fukumizu+etal:2009}.
On the other hand, our proposed approach is equipped with
an objective model selection strategy, which is a notable advantage over CLUHSIC.

Another line of dependence-maximization clustering
adopts \emph{mutual information} (MI) as a dependency measure.
Recently, a dependence-maximization clustering method
called \emph{mean nearest-neighbor} (MNN) clustering
was proposed \citep{ICML:Faivishevsky+Goldberger:2010}.
MNN is based on the $k$-nearest-neighbor entropy estimator
proposed by \citet{PIT:Kozachenko+Leonenko:1987}.

The performance of the original $k$-nearest-neighbor entropy estimator
depends on the choice of the number of nearest neighbors, $k$.
On the other hand, MNN avoids this problem
by introducing a heuristic of taking an average over all possible $k$.
The resulting objective function is given by
\begin{align}
  \sum_{y=1}^{\nclass}\frac{1}{\nsample_y-1}
  \sum_{i\neq j:y_i=y_j=y}\log(\|\boldx_i-\boldx_j\|^2+\epsilon),
  \label{MNN-objective}
\end{align}
where $\epsilon$ ($>0$) is a smoothing parameter.
Then this objective function is minimized with respect to
cluster assignments $\{y_i\}_{i=1}^\nsample$ using a greedy algorithm.

Although the fact that the tuning parameter $k$ is averaged out is convenient,
this heuristic is not well justified theoretically.
Moreover, the choice of the smoothing parameter $\epsilon$ is arbitrary.
In the MATLAB code provided by one of the authors,
$\epsilon=1/\nsample$ was recommended,
but there seems no justification for this choice.
Also, due to the greedy optimization scheme, 
MNN is computationally expensive.
On the other hand, our proposed approach offers
a well-justified model selection strategy,
and the SMI-based clustering gives an analytic-form solution
which can be computed efficiently.

\subsection{Information-Maximization Clustering with Mutual Information}
\label{subsec:MIclustering}
Finally, we review methods of information-maximization clustering
based on \emph{mutual information} \citep{NIPS2005_569,NIPS2010_0457},
which belong to the same family of clustering algorithms as our proposed method.



Mutual information (MI) is defined and expressed as
\begin{align}
  \mathrm{MI}&:=\int\sum_{y=1}^{\nclass} \density(\boldx,y)
  \log\frac{\density(\boldx,y)}{\density(\boldx)\density(y)}
  \mathrm{d}\boldx\nonumber\\
  &\phantom{:}=\int\sum_{y=1}^{\nclass} \density(y|\boldx)\density(\boldx)
  \log\density(y|\boldx)\mathrm{d}\boldx
  -\int\sum_{y=1}^{\nclass} \density(y|\boldx)\density(\boldx)\log\density(y)\mathrm{d}\boldx.
  \label{MI2}
\end{align}
Let us approximate the class-posterior probability $\density(y|\boldx)$ by
a conditional-probability model $\densitymodell(y|\boldx;\boldalpha)$
with parameter $\boldalpha$.
Then the marginal probability $\density(y)$ can be approximated as
\begin{align}
  \density(y)=\int\density(y|\boldx)\density(\boldx)\mathrm{d}\boldx
  \approx\frac{1}{\nsample}\sum_{i=1}^{\nsample}\densitymodell(y|\boldx_i;\boldalpha).
\end{align}
By further approximating the expectation with respect to $\density(\boldx)$
included in Eq.\eqref{MI2}
by the empirical average of samples $\{\boldx_i\}_{i=1}^{\nsample}$,
the following MI estimator can be obtained \citep{NIPS2005_569,NIPS2010_0457}:
\begin{align}
  \widehat{\mathrm{MI}}&:=
  \frac{1}{\nsample}\sum_{i=1}^{\nsample}\sum_{y=1}^{\nclass}
  \densitymodell(y|\boldx_i;\boldalpha)\log\densitymodell(y|\boldx_i;\boldalpha)\nonumber\\
  &\phantom{:=}-\sum_{y=1}^{\nclass}
  \left(\frac{1}{\nsample}\sum_{i=1}^{\nsample}\densitymodell(y|\boldx_i;\boldalpha)\right)
  \log\left(\frac{1}{\nsample}\sum_{j=1}^{\nsample}\densitymodell(y|\boldx_j;\boldalpha)\right).
  \label{MIh}
\end{align}
In \citet{NIPS2005_569},
the Gaussian model,
\begin{align*}
    \densitymodell(y|\boldx;\boldalpha)&
    \propto\exp\left(-\frac{\|\boldx-\boldc_y\|^2}{2s_y^2}+b_y\right),
\end{align*}
(or its kernelized version) is adopted,
where $\boldalpha=\{\boldc_y,s_y,b_y\}_{y=1}^\nclass$ is the parameter.
Then a local maximizer of $\widehat{\mathrm{MI}}$
with respect to the parameter $\boldalpha$ is found
by a gradient method.
On the other hand, in \citet{NIPS2010_0457},
the logistic model
\begin{align}
  \densitymodell(y|\boldx;\boldalpha)&
  \propto\exp\left(\boldalpha_y^\top\boldx\right),
  \label{logistic-model}
\end{align}
(or its kernelized version) is adopted,
where $\boldalpha=\{\boldalpha_y\}_{y=1}^\nclass$ is the parameter.
Then a local maximizer of $\widehat{\mathrm{MI}}$
with respect to the parameter $\boldalpha$ is found
by a quasi-Newton method.


Finally, cluster assignments $\{y_i\}_{i=1}^\nsample$ are determined as
\begin{align*}
  y_i&=\argmax_y \densitymodell(y|\boldx_i;\boldalphah),
\end{align*}
where $\boldalphah$ is a local maximizer of $\widehat{\mathrm{MI}}$.
Below, we refer to the above method as \emph{MI-based clustering} (MIC).

In the kernelized version of MIC,
the user needs to determine 
parameters included in the kernel function
such as the kernel width or the number of nearest neighbors.
\citet{NIPS2005_569} proposed to choose the
kernel parameters so that $\widehat{\mathrm{MI}}$ \eqref{MIh} is maximized.
Thus, cluster assignments and kernel parameters can be consistently determined 
under the common guidance of maximizing $\widehat{\mathrm{MI}}$.
However, since $\widehat{\mathrm{MI}}$ is an unsupervised estimator of MI,
it is not accurately enough;
in the model selection stage, 
cluster labels $\{y_i\}_{i=1}^\nsample$ are available and thus supervised
estimation of MI is more favorable.
Indeed, there exists a more powerful supervised MI estimator
called \emph{maximum-likelihood MI} \citep[MLMI;][]{FSDM:Suzuki+etal:2008},
which was proved to achieve the optimal non-parametric convergence rate.

The derivation of MLMI follows a similar line to LSMI explained in Section~\ref{subsec:LSMI},
i.e., the density-ratio function \eqref{density-ratio} is learned.
More specifically, the following density-ratio model $\ratiomodel(\boldx,y;\boldtheta)$
is used:
\begin{align*}
\ratiomodel(\boldx,y;\boldtheta) &:=\sum_{\ell:y_i=y}\theta_{\ell}L_{}(\boldx,\boldx_\ell),
\end{align*}
where $\boldtheta=(\theta_1,\ldots,\theta_{\nsample})^\top$
and
$L_{}(\boldx,\boldx')$ is a kernel function
with a kernel parameter $\kernelparameterr$.
Then the parameter $\boldtheta$ is learned so that 
the Kullback-Leibler divergence from $\density(\boldx,y)$ to
$\ratiomodel(\boldx,y;\boldtheta)\density(\boldx)\density(y)$
is minimized\footnote{
Note that $\ratiomodel(\boldx,y;\boldtheta)\density(\boldx)\density(y)$
can be regarded as a model of $\density(\boldx,y)$.}.
An empirical version of the MLMI optimization problem is given as
\begin{align*}
  \max_{\boldtheta}\;\;&\frac{1}{\nsample}\sum_{i=1}^{\nsample} \log\ratiomodel(\boldx_i,y_i;\boldtheta)\\
  \mbox{s.t.~}\;\; &\frac{1}{\nsample^2}\sum_{i,j=1}^{\nsample} \ratiomodel(\boldx_i,y_j;\boldtheta)=1
  \mbox{~~~and~~~} \boldtheta\ge \boldzero_{\nsample},
\end{align*}
where
$\boldzero_{\nsample}$ denotes the $\nsample$-dimensional vector with all zeros
and the inequality for vectors is applied in the element-wise manner.
This is a convex optimization problem,
and thus the global optimal solution $\boldthetah$,
which tends to be sparse, can be easily obtained by, e.g.,
a projected gradient method \citep{AISM:Sugiyama+etal:2008}.

Then an MI estimator called MLMI is given as follows:
\begin{align*}
  \mathrm{MLMI}
  := \frac{1}{\nsample} \sum_{i=1}^{\nsample} \log
\ratiomodel(\boldx_i,y_i;\boldthetah).
\end{align*}
The kernel parameter $\kernelparameterr$ included in the kernel function 
$L_{}(\boldx,\boldx')$ can be optimized by cross-validation,
in the same way as LSMI \citep{FSDM:Suzuki+etal:2008}.


\section{Experiments}
\label{sec:experiments}
In this section, we experimentally evaluate the performance of the proposed
and existing clustering methods.

\subsection{Illustration}
\label{subsec:illustration}
\begin{figure}[p]
  \centering
  \footnotesize
  \begin{tabular}{@{}cccc@{}}
    $\mathrm{ARI}=1$ &
    $\mathrm{ARI}=1$ &
    $\mathrm{ARI}=1$ &
    $\mathrm{ARI}=0.773$\\
    \includegraphics[width=0.22\textwidth,clip]{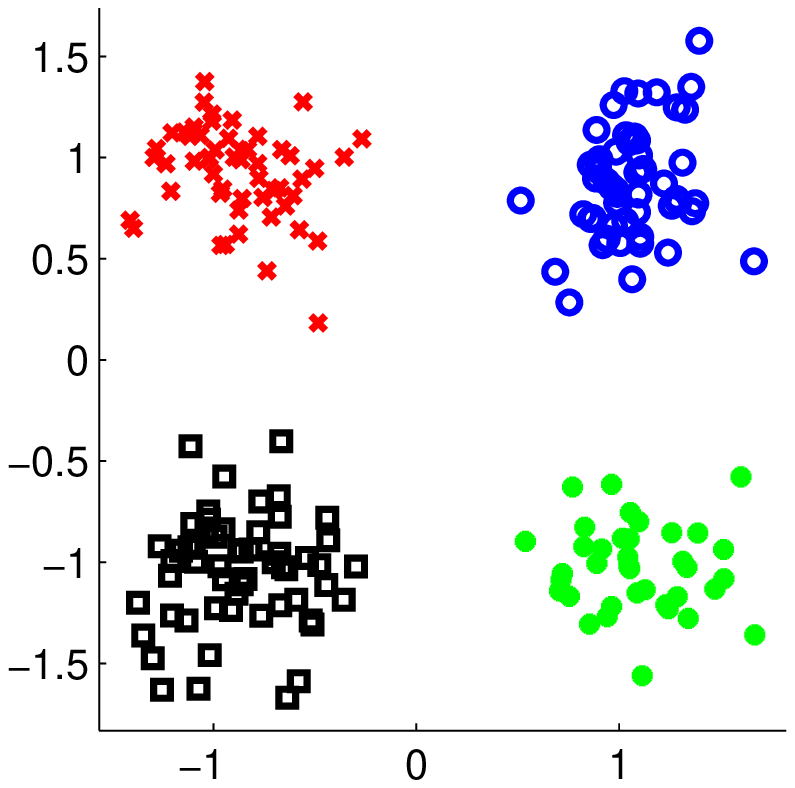} &
    \includegraphics[width=0.22\textwidth,clip]{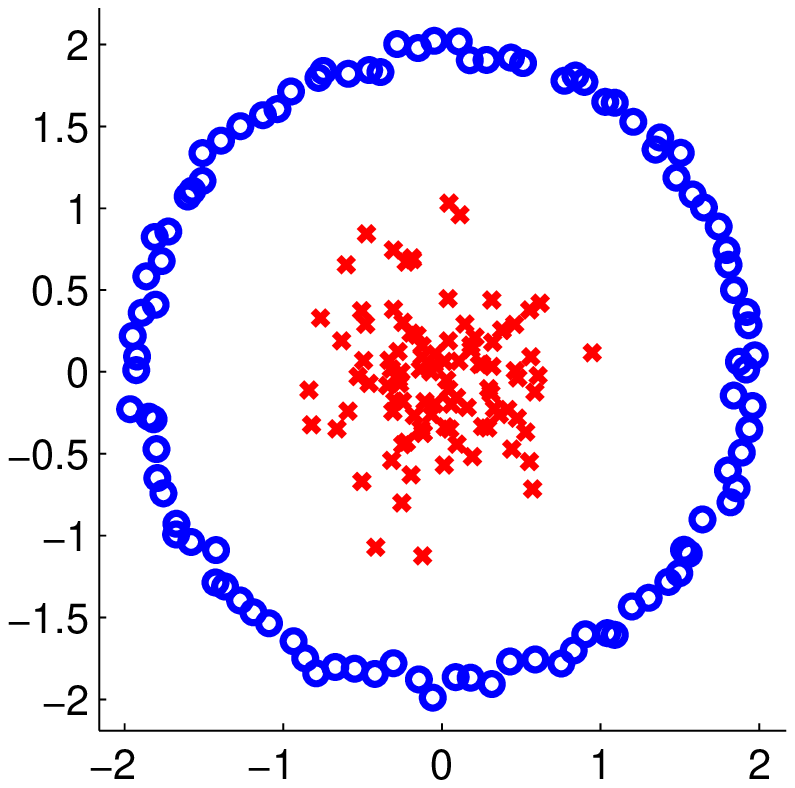} &
    \includegraphics[width=0.22\textwidth,clip]{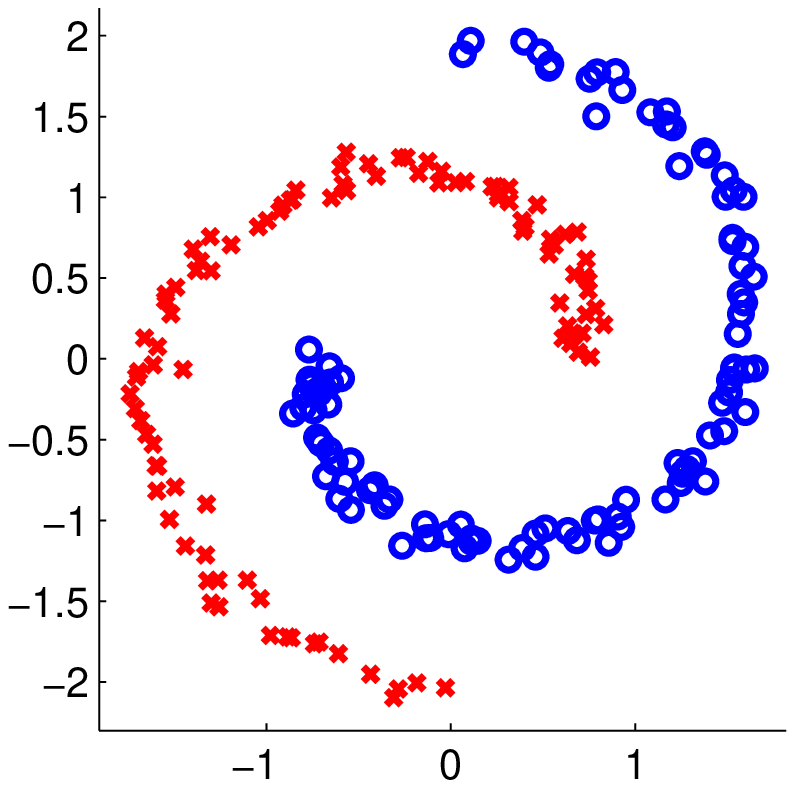} &
    \includegraphics[width=0.22\textwidth,clip]{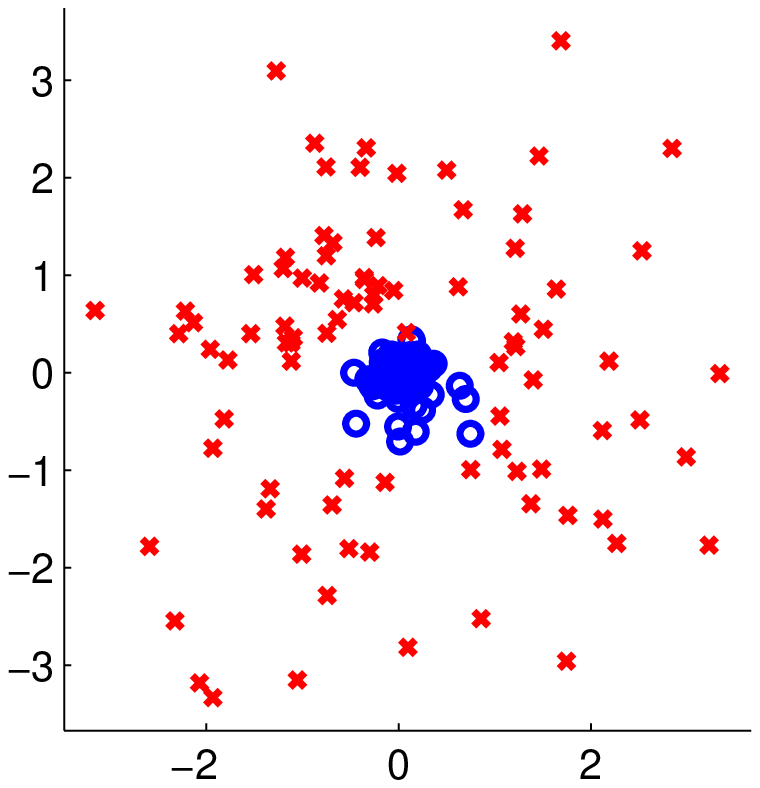} \\
    \includegraphics[width=0.22\textwidth,clip]{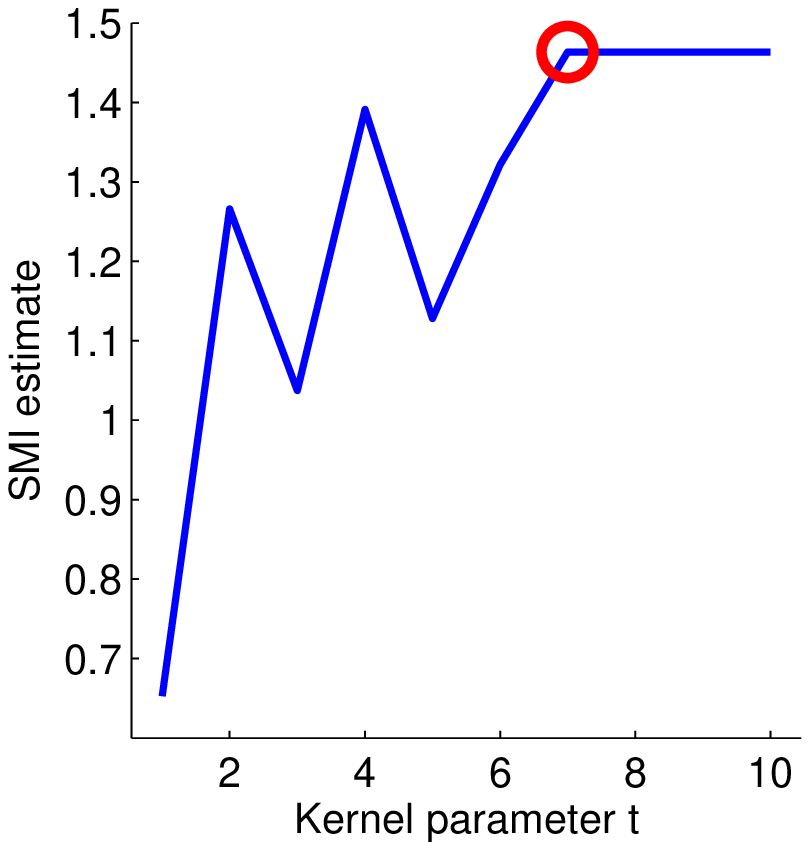} &
    \includegraphics[width=0.22\textwidth,clip]{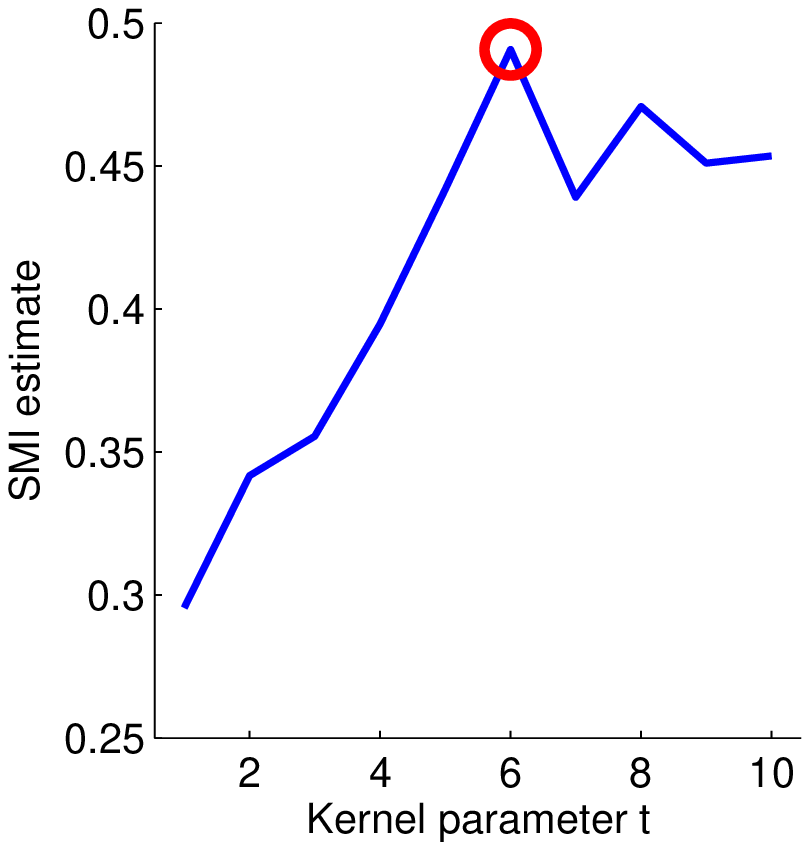} &
    \includegraphics[width=0.22\textwidth,clip]{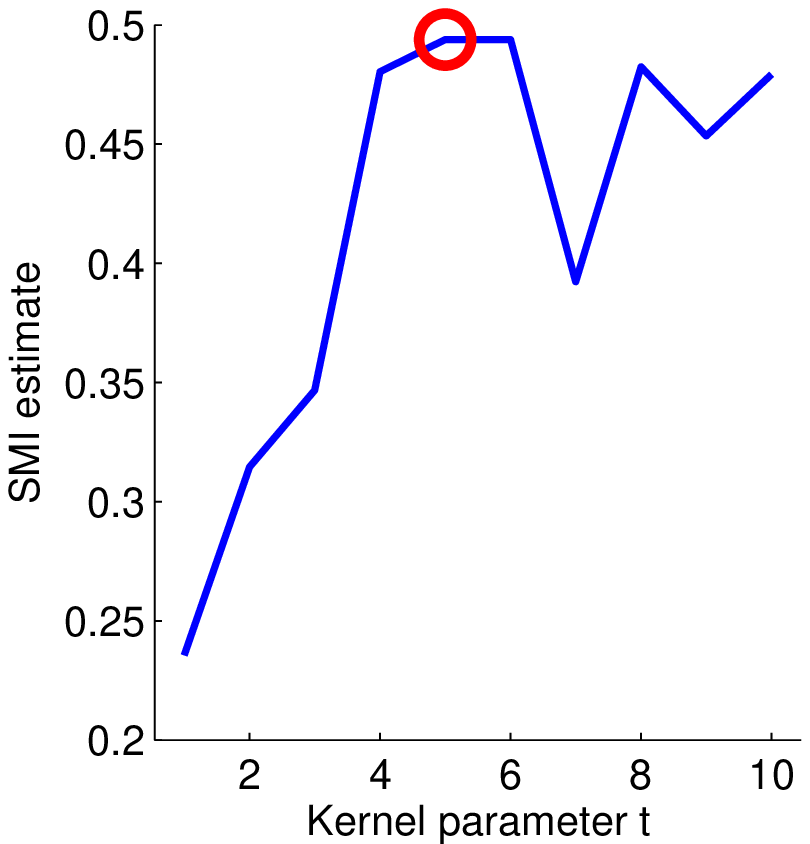} &
    \includegraphics[width=0.22\textwidth,clip]{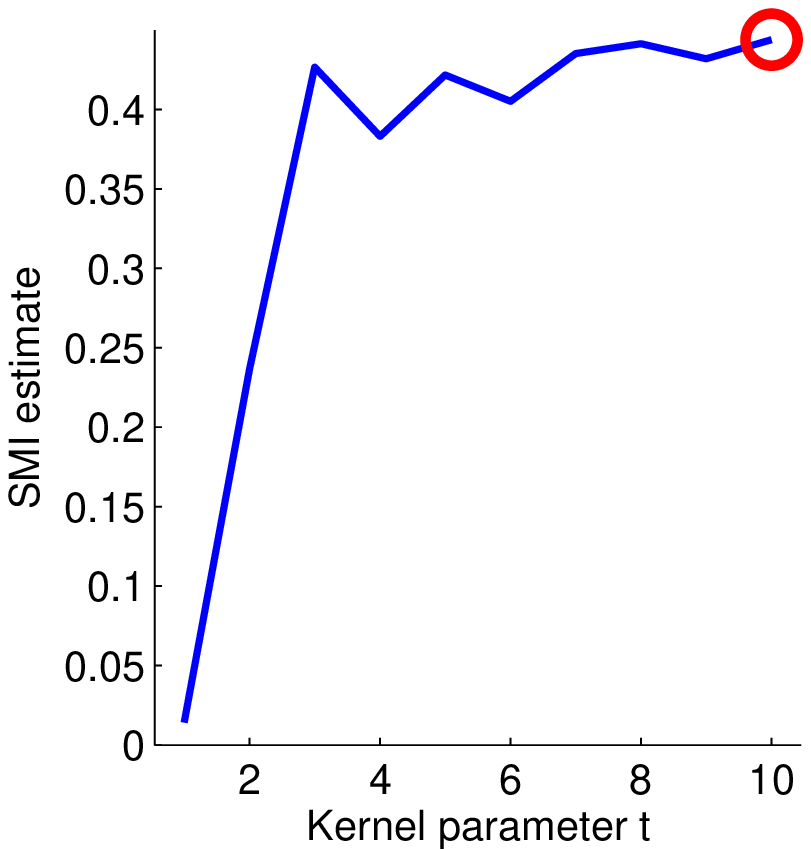} \\
    (a) Four Gaussian blobs &
    (b) Circle \& Gaussian &
    (c) Double spirals &
    (d) High \& low densities\\
  \end{tabular}
  \caption{Illustrative examples. Cluster assignments obtained by SMIC (top)
    and model selection curves obtained by LSMI (bottom).}
  \label{fig:toy-SMIC}
\vspace*{10mm}
  \begin{tabular}{@{}cccc@{}}
    $\mathrm{ARI}=1$ &
    $\mathrm{ARI}=1$ &
    $\mathrm{ARI}=0.021$ &
    $\mathrm{ARI}=0.173$\\
    \includegraphics[width=0.22\textwidth,clip]{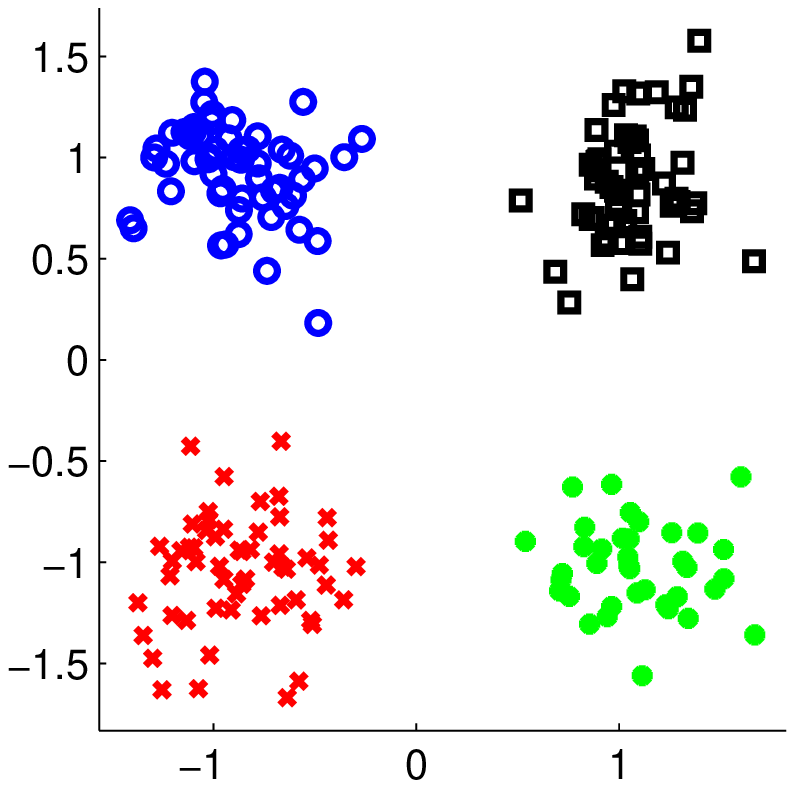} &
    \includegraphics[width=0.22\textwidth,clip]{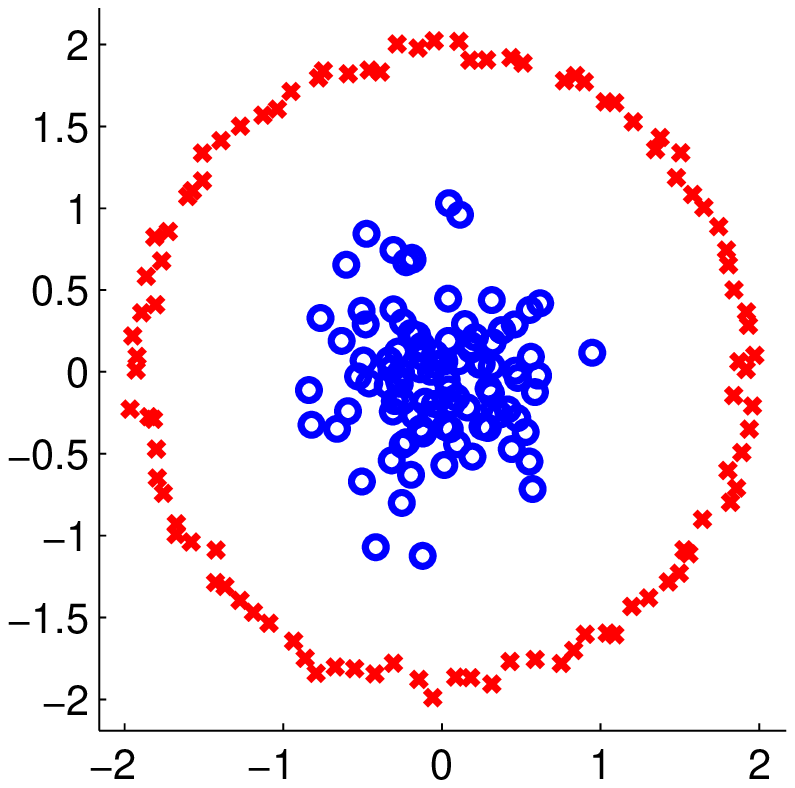} &
    \includegraphics[width=0.22\textwidth,clip]{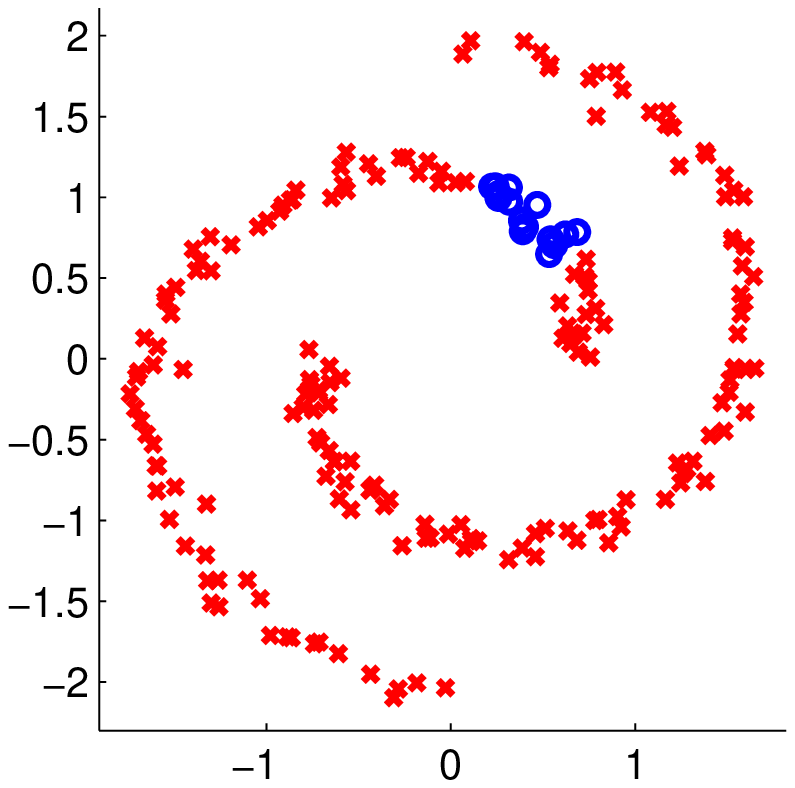} &
    \includegraphics[width=0.22\textwidth,clip]{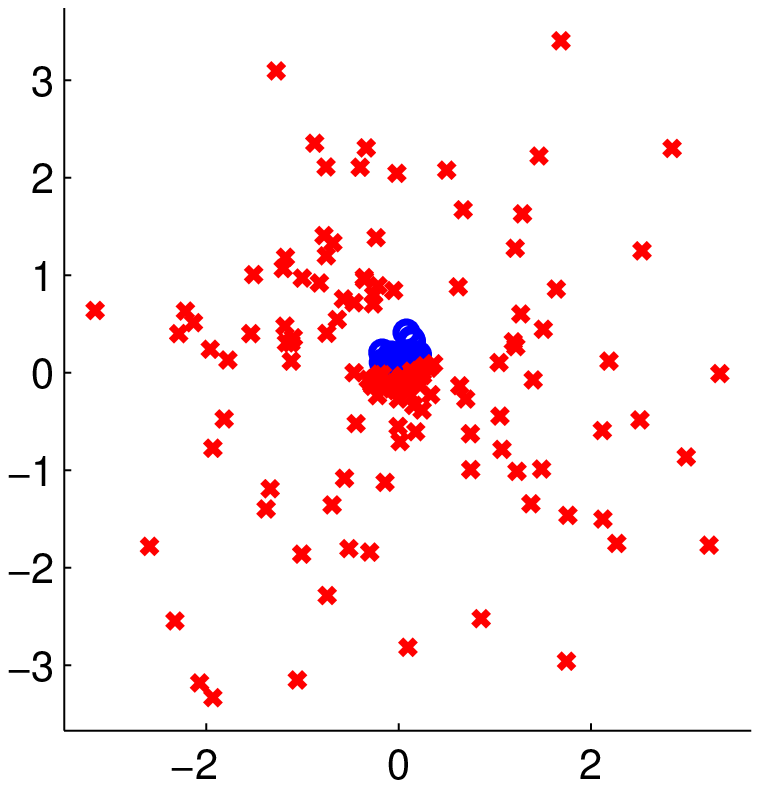} \\
    \includegraphics[width=0.22\textwidth,clip]{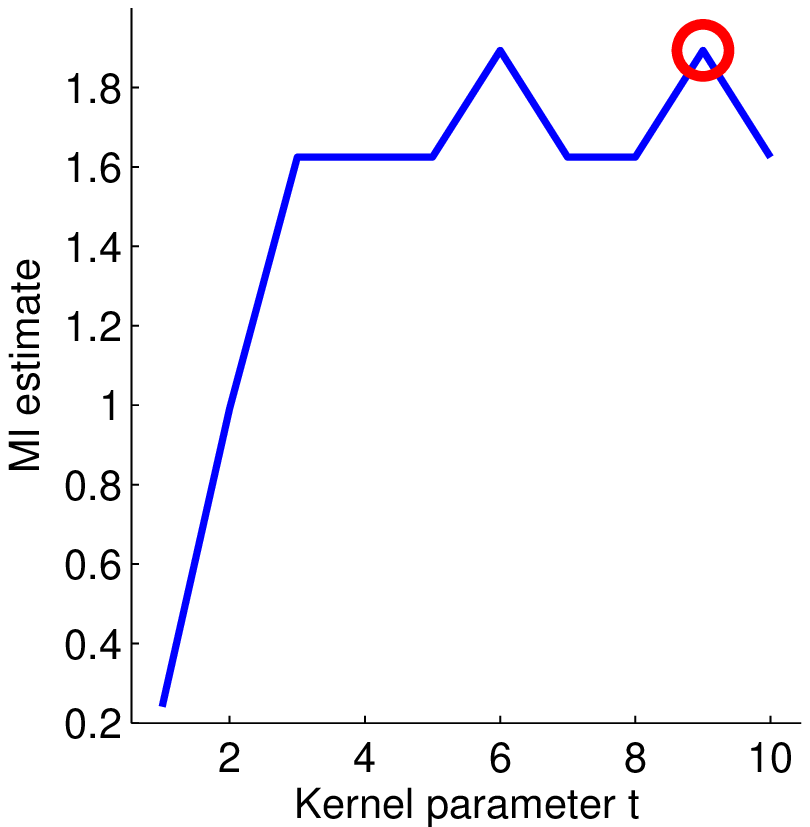} &
    \includegraphics[width=0.22\textwidth,clip]{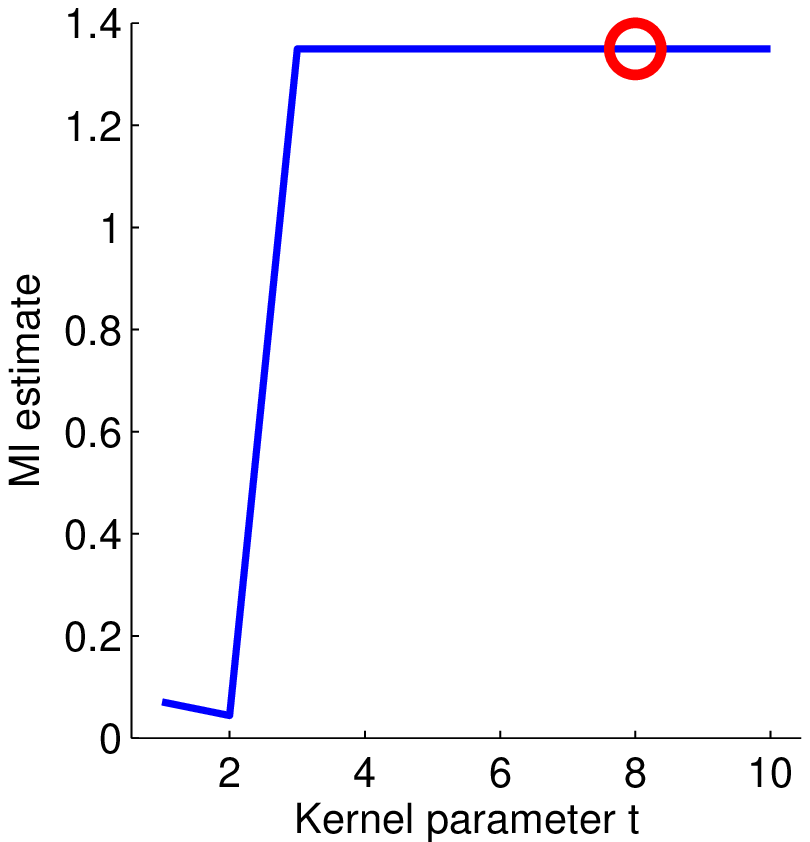} &
    \includegraphics[width=0.22\textwidth,clip]{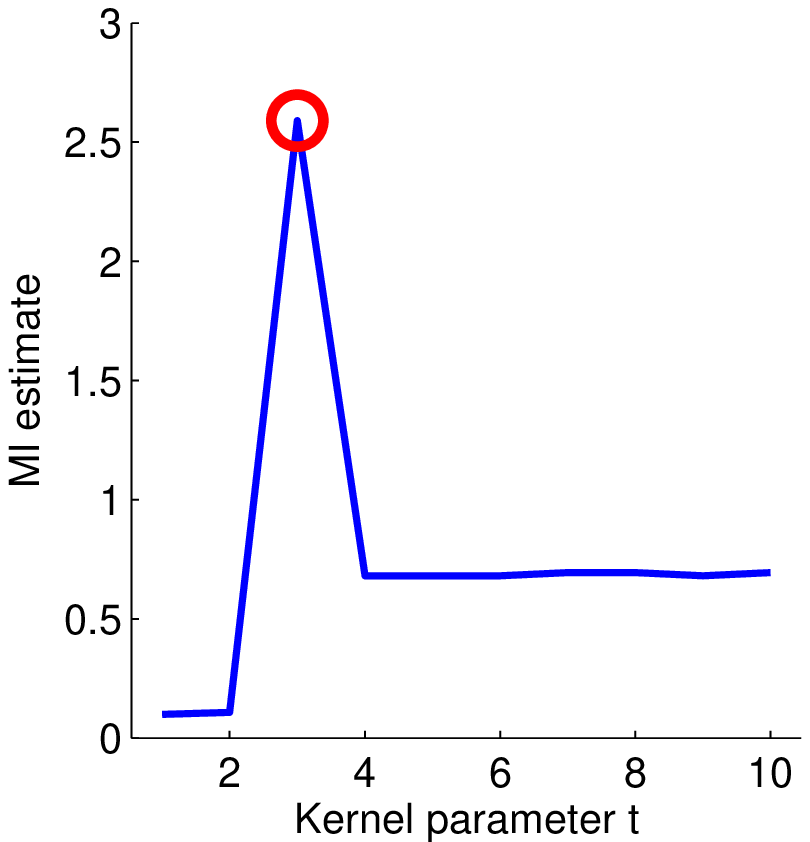} &
    \includegraphics[width=0.22\textwidth,clip]{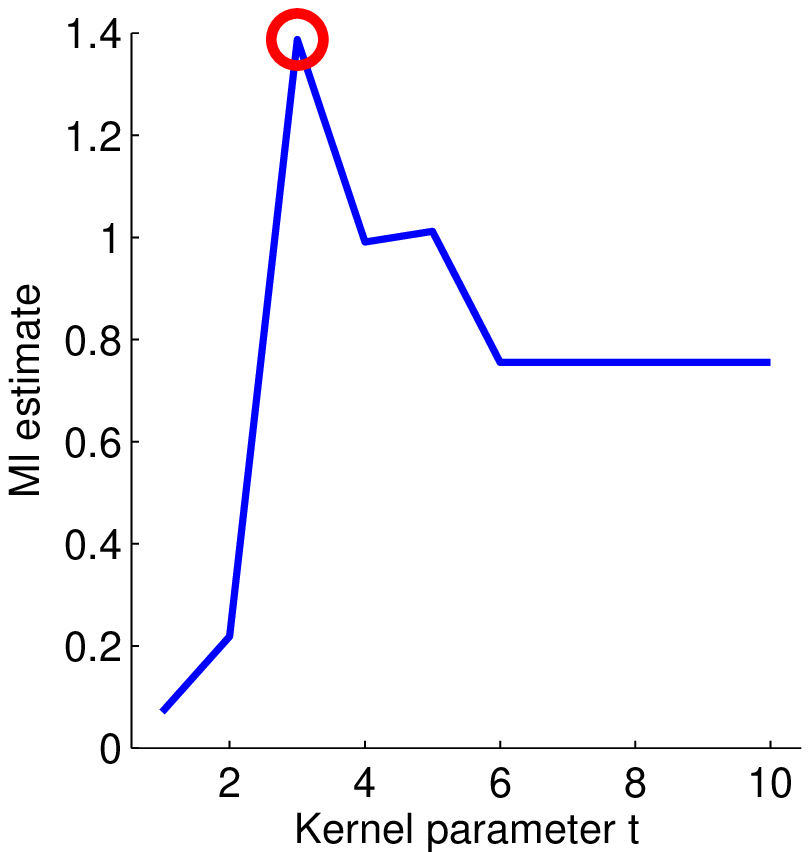} \\
    (a) Four Gaussian blobs &
    (b) Circle \& Gaussian &
    (c) Double spirals &
    (d) High \& low densities\\
  \end{tabular}
  \caption{Illustrative examples. Cluster assignments obtained by MIC (top)
    and model selection curves obtained by MLMI (bottom).}
  \label{fig:toy-MIC}
 \end{figure}

First, we illustrate the behavior of the proposed method
using the following $4$ artificial datasets
with dimensionality $\inputdim=2$ and sample size $\nsample=200$:
\begin{description}
\item[(a) Four Gaussian blobs:] For the number of classes $\nclass=4$, samples
  in each class are drawn from
  the Gaussian distributions with
  mean $(2,2)^\top$, $(-2,2)^\top$, $(2,-2)^\top$, and $(-2,-2)^\top$
  and covariance matrix $0.25\boldI_2$, respectively.
\item[(b) Circle \& Gaussian:]
  For $\nclass=2$, 
  samples in one class are drawn from the $2$-dimensional standard normal distribution,
  and samples in the other class are equi-distantly located
  on the origin-centered circle with radius $5$.
  Then noise following the origin-centered normal distribution
  with covariance matrix $0.01\boldI_2$ is added to each sample.
\item[(c) Double spirals:] For $\nclass=2$,
  the $i$-th sample in one class is given by 
  $\left(\ell_i\cos(m_i),\ell_i\sin(m_i)\right)^\top$,
  and
  the $i$-th sample in the other class is given by 
  $\left(-\ell_i\cos(m_i),-\ell_i\sin(m_i)\right)^\top$,
  where $\ell_i=1+4(i-1)/\nsample$ and $m_i=3\pi(i-1)/\nsample$.
  Then noise following the origin-centered normal distribution
  with covariance matrix $0.01\boldI_2$ is added to each sample.
\item[(d) High \& low densities:] For $\nclass=2$,
  samples in one class are drawn from the $2$-dimensional standard normal distribution,
  and
  samples in the other class are drawn from
  the $2$-dimensional origin-centered normal distribution
  with covariance matrix $0.01\boldI_2$.
\end{description}
The class-prior probability was set to be uniform.
The generated samples were centralized and their variance was normalized
in the dimension-wise manner (see the top row of Figure~\ref{fig:toy-SMIC}).
A MATLAB code for generating these samples are available from
\begin{center}
  `\url{http://sugiyama-www.cs.titech.ac.jp/~sugi/software/SMIC}'.
\end{center}
As a kernel function,
we used the sparse local-scaling kernel \eqref{sparse-local-scaling} for SMIC,
where the kernel parameter $\kernelparameter$ was chosen from\footnote{
  We confirmed that $\kernelparameter$ larger than $10$ was not chosen in this experiment.}
$\{1,\ldots,10\}$ based on LSMI with the Gaussian kernel \eqref{Gaussian-kernel}.


The top graphs in Figure~\ref{fig:toy-SMIC} depict
the cluster assignments obtained by SMIC with the uniform class-prior,
and the bottom graphs in Figure~\ref{fig:toy-SMIC} depict
the model selection curves obtained by LSMI
(i.e., the values of LSMI as functions of the model parameter $\kernelparameter$).
The clustering performance was evaluated
by the \emph{adjusted Rand index} \citep[ARI;][]{JoC:Hubert+Arabie:1985}
between inferred cluster assignments and the ground truth categories
(see Appendix for the details of ARI).
Larger ARI values mean better performance,
and ARI takes its maximum value $1$ when two sets of cluster assignments are identical.
The results show that SMIC combined with LSMI works well for these toy datasets.

Figure~\ref{fig:toy-MIC} depicts the
cluster assignments and model selection curves
obtained by MIC with MLMI (see Section~\ref{subsec:MIclustering}),
where
pre-training of the kernel logistic model using the cluster assignments obtained by
\emph{self-tuning spectral clustering} \citep{NIPS17:Zelnik-Manor+Perona:2005}
was carried out for initializing MIC
\citep{NIPS2010_0457}.
The figure shows that qualitatively good clustering results were obtained
for the datasets (a) and (b).
However, for the datasets (c) and (d), 
poor results were obtained due to local optima of the objective function \eqref{MIh}.

\begin{figure}[p]
  \centering
  \subfigure[Four Gaussian blobs]{
    \begin{tabular}{@{}c@{}}
      \includegraphics[width=0.45\textwidth,clip]{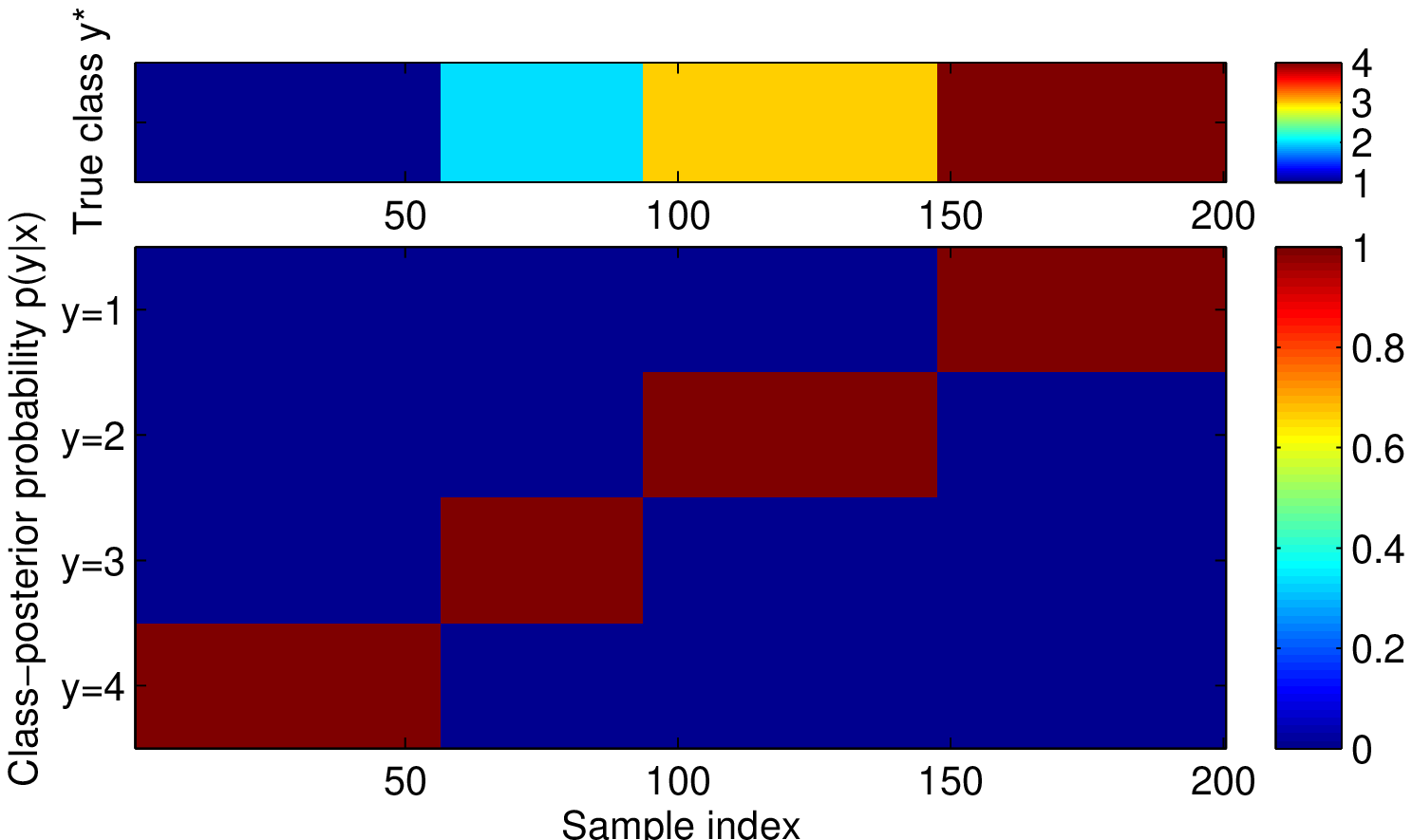}
    \end{tabular}
  }
  \subfigure[Circle \& Gaussian]{
    \begin{tabular}{@{}c@{}}
      \includegraphics[width=0.45\textwidth,clip]{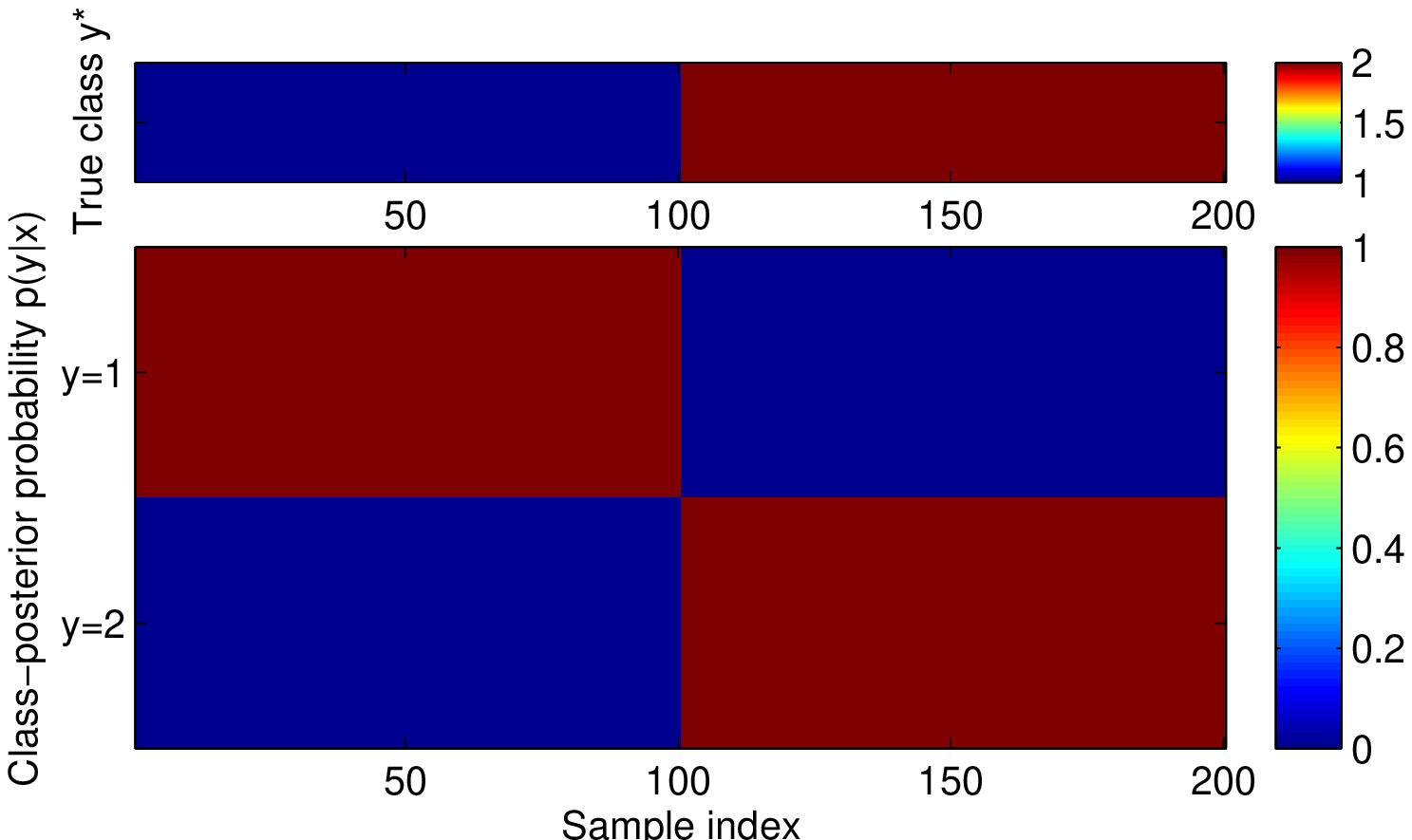}
    \end{tabular}
  }
  \subfigure[Double spirals]{
    \begin{tabular}{@{}c@{}}
      \includegraphics[width=0.45\textwidth,clip]{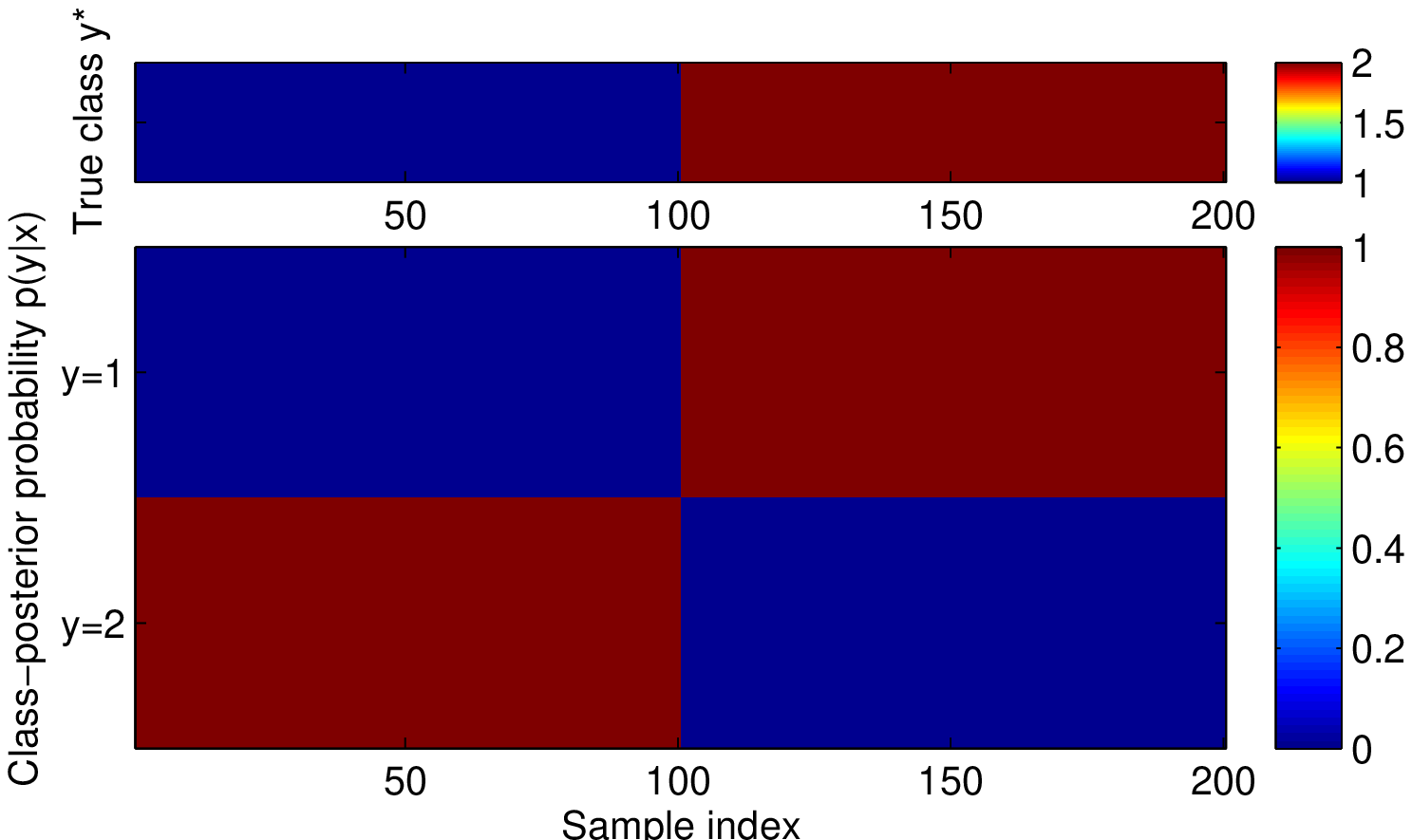}
    \end{tabular}
  }
  \subfigure[High \& low densities]{
    \begin{tabular}{@{}c@{}}
      \includegraphics[width=0.45\textwidth,clip]{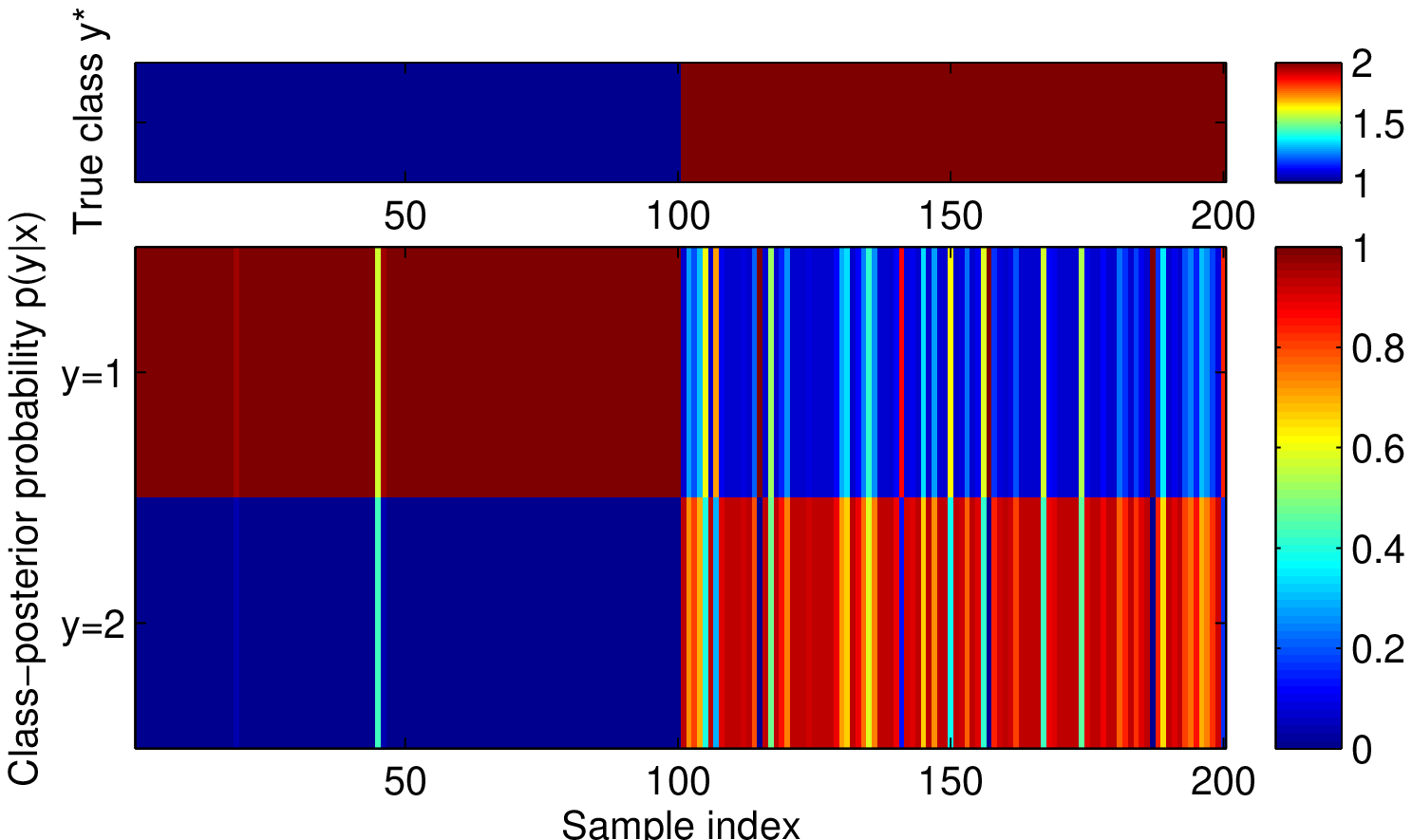}
    \end{tabular}
  }
  \caption{Illustrative examples. Class-posterior probabilities estimated by SMIC.}
  \label{fig:toy-probability-SMIC}
\vspace*{10mm}
  \subfigure[Four Gaussian blobs]{
    \begin{tabular}{@{}c@{}}
      \includegraphics[width=0.45\textwidth,clip]{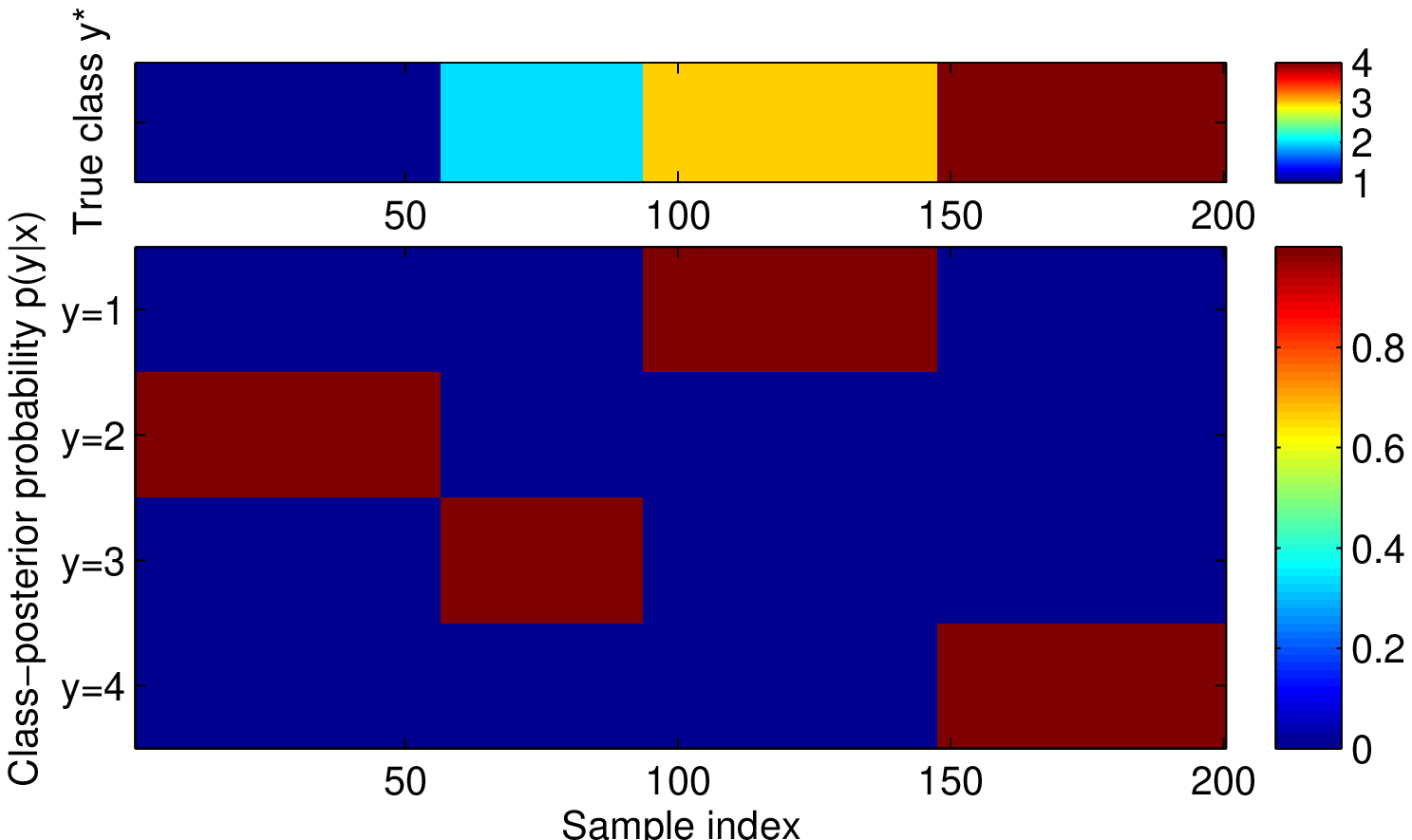}
    \end{tabular}
  }
  \subfigure[Circle \& Gaussian]{
    \begin{tabular}{@{}c@{}}
      \includegraphics[width=0.45\textwidth,clip]{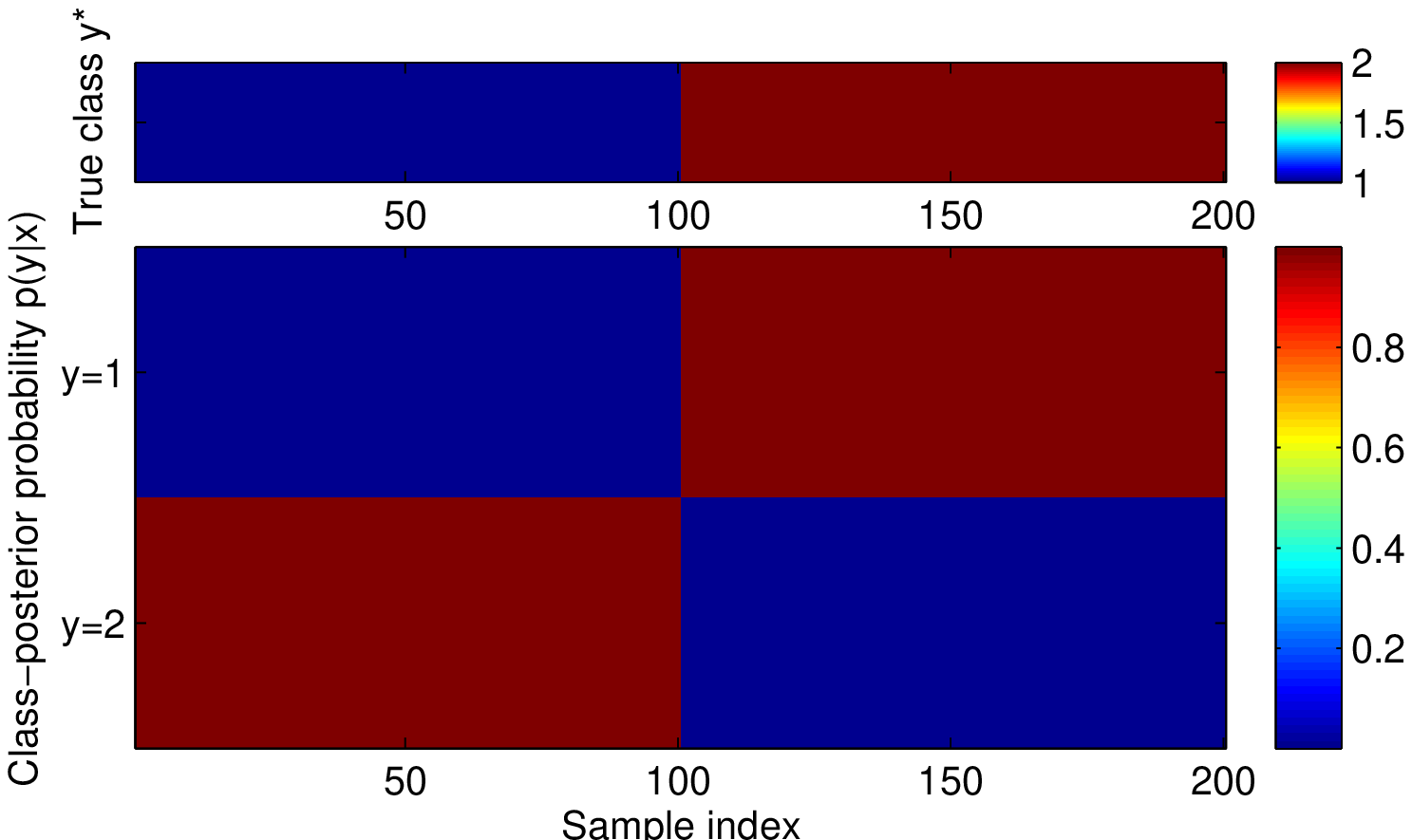}
    \end{tabular}
  }
  \subfigure[Double spirals]{
    \begin{tabular}{@{}c@{}}
      \includegraphics[width=0.45\textwidth,clip]{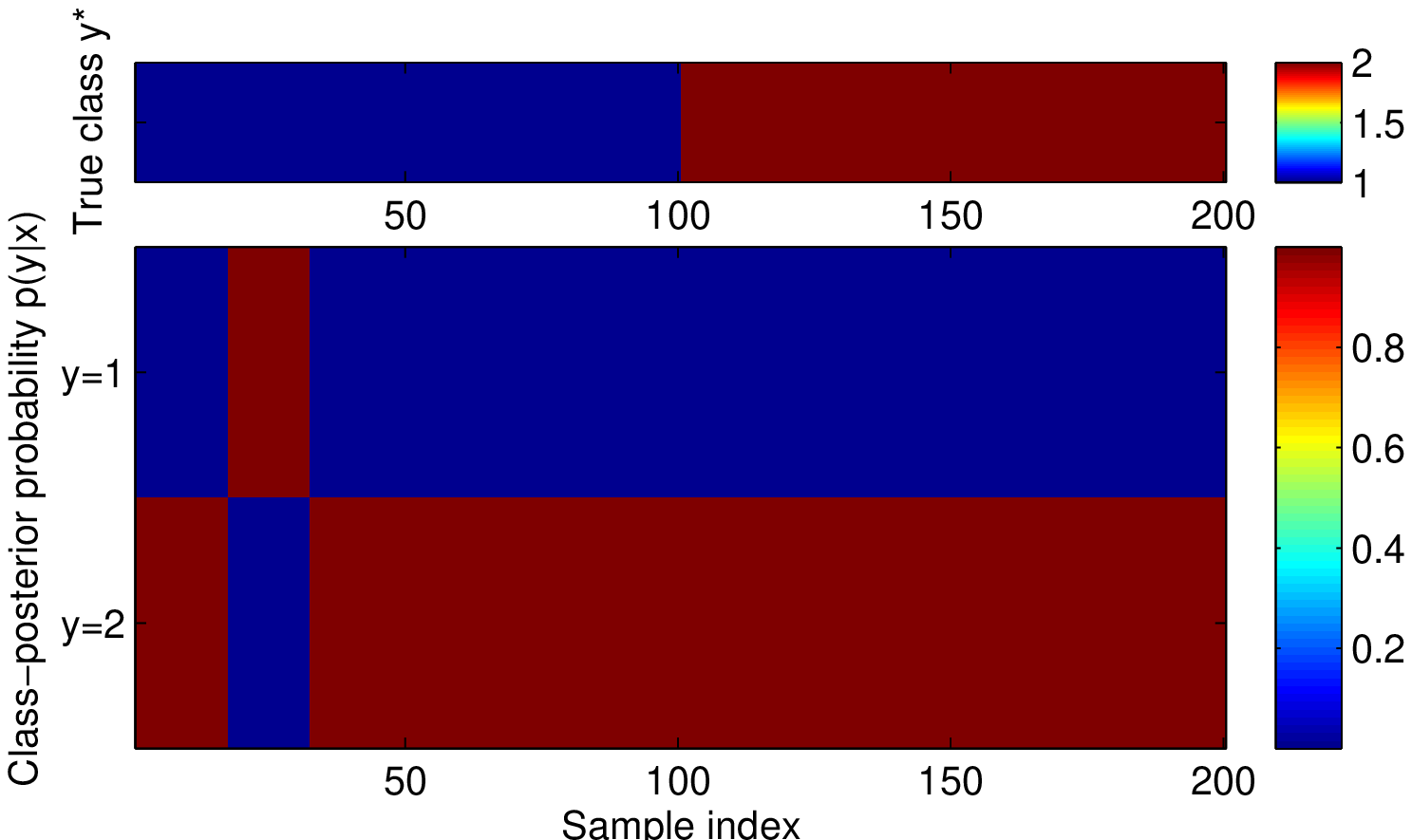}
    \end{tabular}
  }
  \subfigure[High \& low densities]{
    \begin{tabular}{@{}c@{}}
      \includegraphics[width=0.45\textwidth,clip]{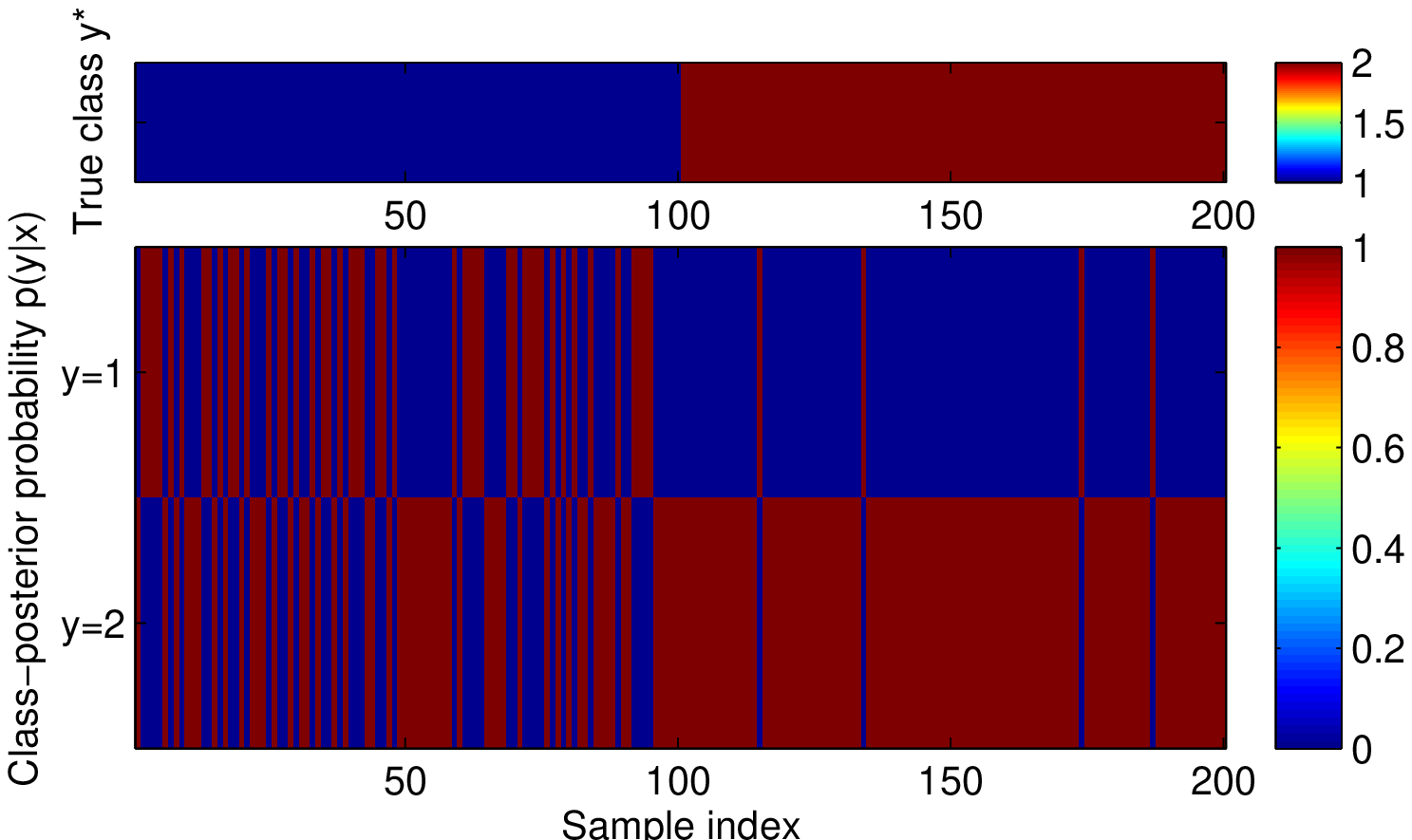}
    \end{tabular}
  }
  \caption{Illustrative examples. Class-posterior probabilities estimated by MIC.}
  \label{fig:toy-probability-MIC}
\end{figure}

Figure~\ref{fig:toy-probability-SMIC}
and
Figure~\ref{fig:toy-probability-MIC}
depict class-posterior probabilities estimated by SMIC and MIC, respectively.
The plots show that, for the datasets (a), (b), and (c) where
the clusters are clearly separated, 
the estimated class-posterior probabilities are almost zero-one functions
and thus the class prediction is highly certain.
On the other hand, for the dataset (d) where
the two clusters are overlapped,
the estimated class-posterior probabilities tend to take
intermediate class-posterior probabilities.


\subsection{Influence of Imbalanced Class-Prior Probabilities}
Next, we experimentally investigate 
how imbalanced class-prior probabilities
(i.e., the sample size in each cluster is significantly different)
influence the clustering performance of SMIC.

We continue using the $4$ artificial datasets 
used in Section~\ref{subsec:illustration},
but we set the true class-prior probability as
\begin{align*}
  \density(y=1)&=\density(y=2)=0.1,0.15,0.2,0.25,\\
  \density(y=3)&=\density(y=4)=\frac{1-\density(y=1)-\density(y=2)}{2},
\end{align*}
for the dataset (a), and
\begin{align*}
\density(y=1)&=0.2,0.3,0.4,0.5,\\
\density(y=2)&=1-\density(y=1),
\end{align*}
for the datasets (b)--(d).
The following $2$ approaches are compared:
\begin{description}
\item[SMIC:] 
  SMIC with the uniform class-prior probabilities $\pi_1=\pi_2=1/2$.
\item[SMIC$^*$:] 
  SMIC with the true class-prior probabilities $\pi_1=\density(y=1)$ and $\pi_2=\density(y=2)$.
\end{description}

\begin{figure}[t]
  \centering
  \subfigure[Four Gaussian blobs]{
    \includegraphics[width=0.45\textwidth,clip]{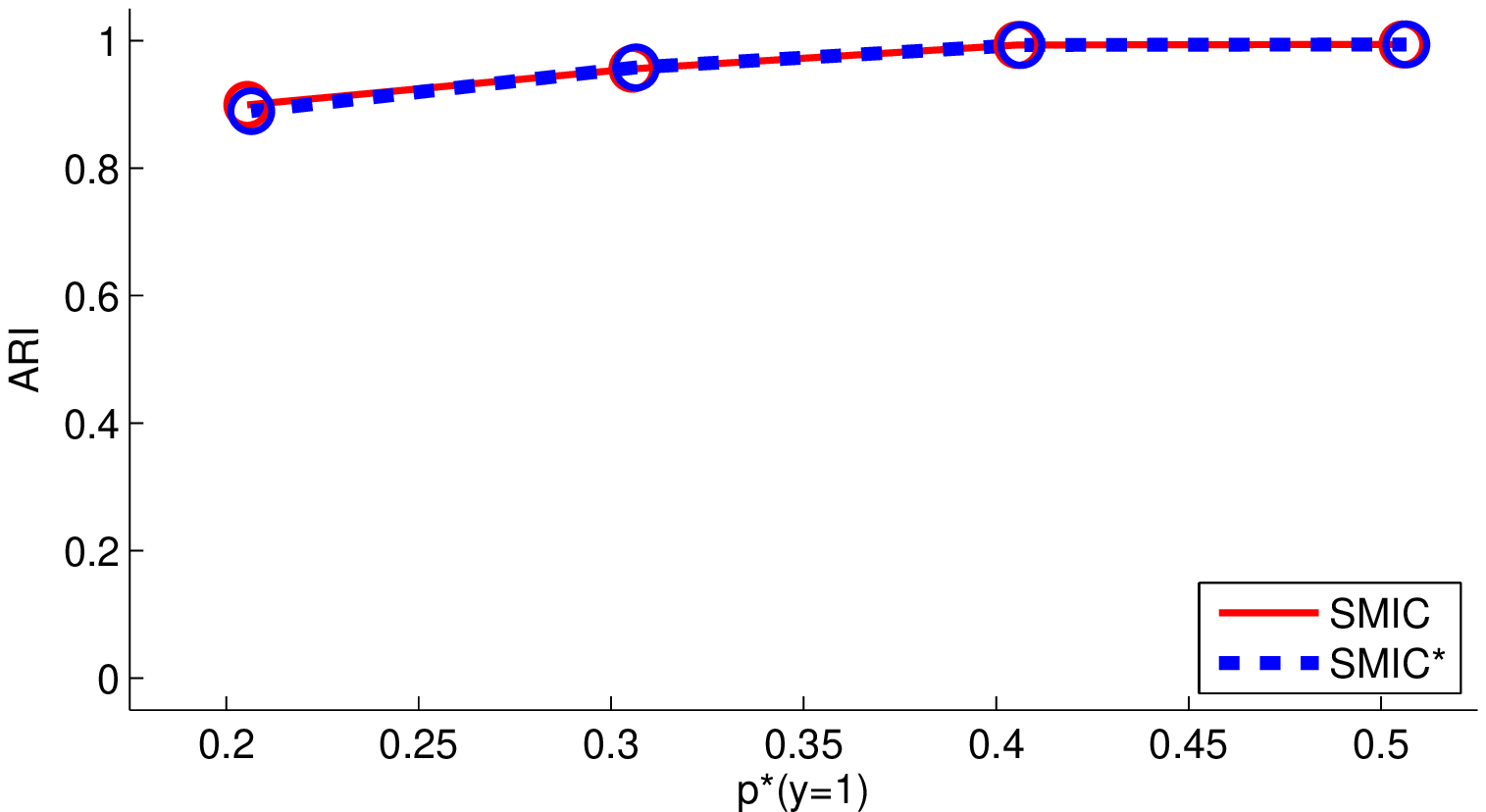}
  }
  \subfigure[Circle and Gaussian]{
    \includegraphics[width=0.45\textwidth,clip]{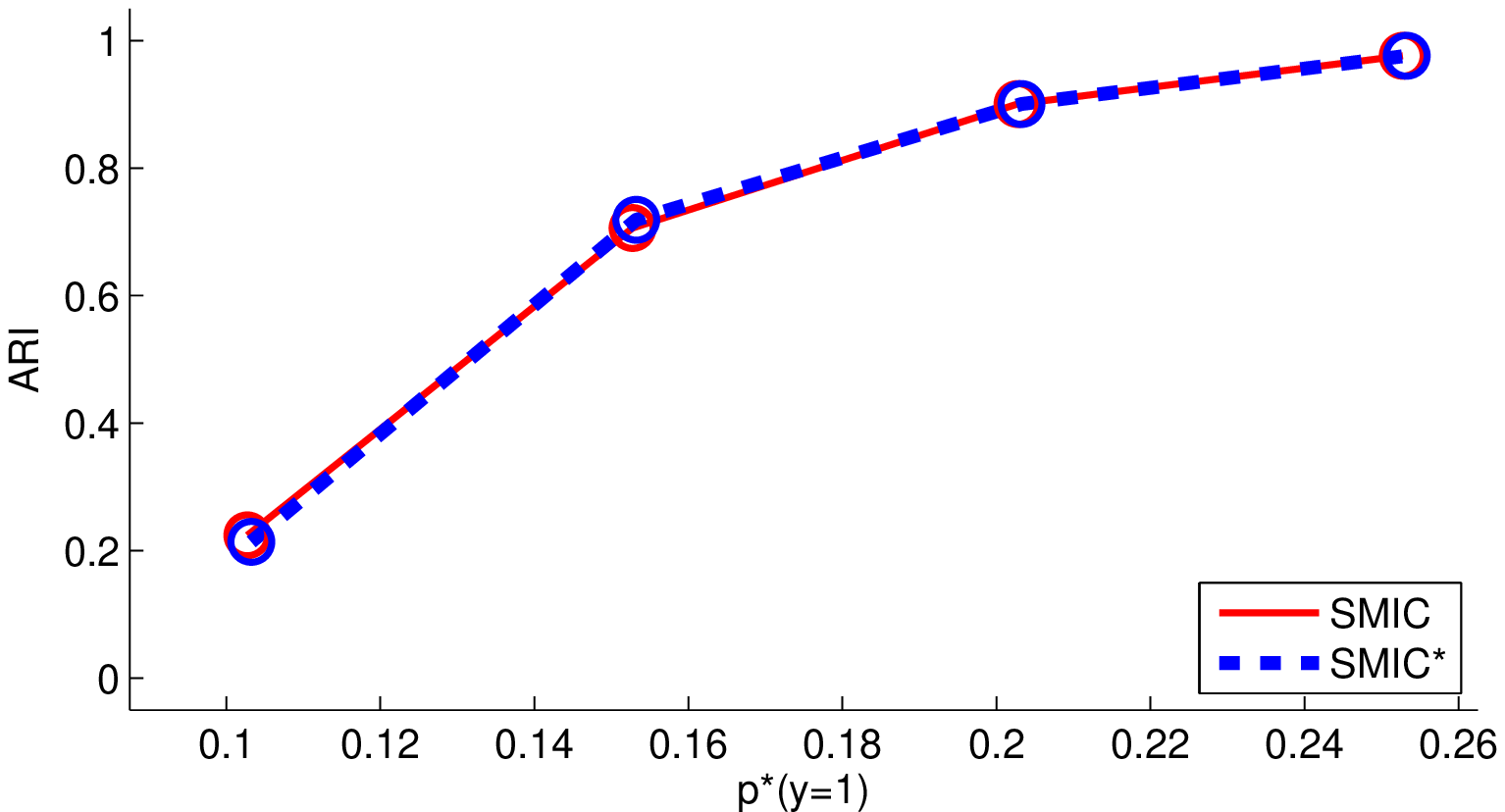}
  }
  \subfigure[Double spirals]{
    \includegraphics[width=0.45\textwidth,clip]{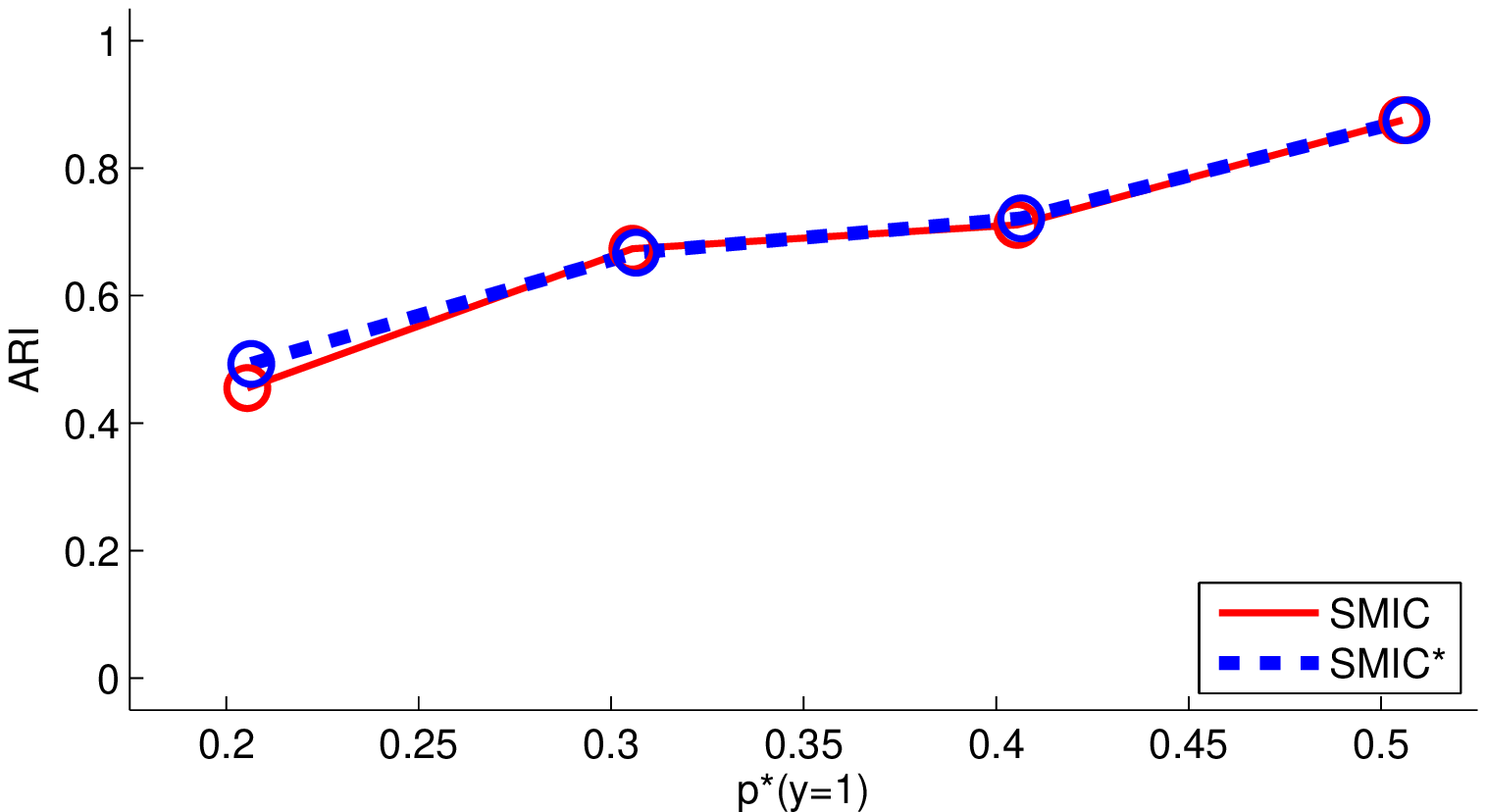}
  }
  \subfigure[High and low densities]{
    \includegraphics[width=0.45\textwidth,clip]{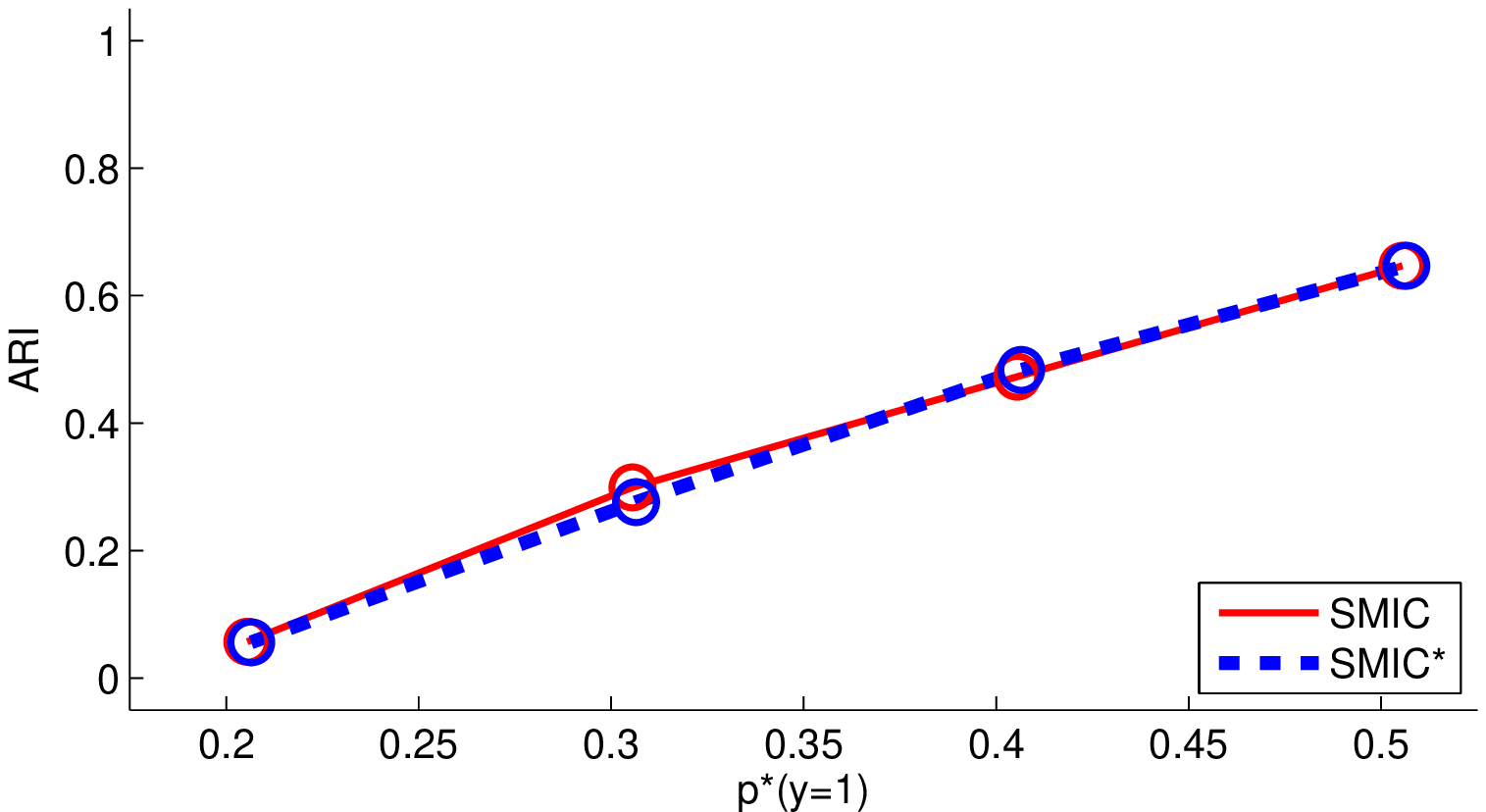}
  }
  \caption{Illustrative examples. The mean ARI over $100$ runs
    as functions of the class-prior probability $\density(y=1)$.
    The two methods were judged to be comparable
    in terms of the average ARI
    by the \emph{t-test} at the significance level $1\%$.
}
  \label{fig:prior}
\end{figure}

The mean and standard deviation of ARI over $100$ runs
are plotted in Figure~\ref{fig:prior},
showing that the difference between SMIC and SMIC$^*$ is negligibly small.
Indeed, the two methods were judged to be comparable to each other
in terms of the average ARI
by the \emph{t-test} at the significance level $1\%$
for all tested cases.
This implies that SMIC is not sensitive to the choice of class-prior probabilities.
Thus, in practice, SMIC with the uniform class-prior distribution may be used
when the true class-prior is unknown.

\subsection{Performance Comparison}
Finally, we systematically compare the performance of the proposed
and existing clustering methods
using various real-world datasets
such as images, natural languages, accelerometric sensors, and speech.

\subsubsection{Setup}
We compared the performance of the following methods,
which all do not contain open tuning parameters
and therefore experimental results are fair and objective:
\begin{description}
\item[KM:] K-means \citep[][see also Section~\ref{subsec:k-means}]{BerkeleySymp:MacQueen:1967}.
  We used the software included in the MATLAB Statistics Toolbox,
  where initial values were randomly generated $100$ times
  and the best result in terms of the k-means objective value was chosen
  as the final solution.

\item[SC:] Spectral clustering \citep[][see also Section~\ref{subsec:spectral-clustering}]{IEEE-PAMI:Shi+Malik:2000,nips02-AA35}
  with the self-tuning local-scaling similarity \citep{NIPS17:Zelnik-Manor+Perona:2005}.
  We used the MATLAB code provided by one of the authors\footnote{
    \url{http://webee.technion.ac.il/~lihi/Demos/SelfTuningClustering.html}},
  where the post k-means processing was repeated $10$ times with heuristic initialization:
  the first center was chosen randomly from samples,
  and then the next center was iteratively set to the farthest sample from the previous ones.
  The best result in terms of the k-means objective value
  out of $10$ repetitions was chosen as the final solution.

\item[MNN:] Mean nearest-neighbor clustering \citep[][see also Section~\ref{subsec:dependence-maximization}]{ICML:Faivishevsky+Goldberger:2010}.
  We used the MATLAB code provided by one of the authors\footnote{
    \url{http://www.levfaivishevsky.webs.com/NIC.rar}}.
  Following the suggestions provided in the program code,
  the number of iterations was set to $10$ 
  and the smoothing parameter $\epsilon$ (see Eq.\eqref{MNN-objective})
  was set to $\epsilon=1/\nsample$.

\item[MIC:] MI-based clustering with kernel logistic models
  and the sparse local-scaling kernel \citep[][see also Section~\ref{subsec:MIclustering}]{NIPS2010_0457},
  where model selection is carried out by 
  maximum-likelihood MI \citep[MLMI;][]{FSDM:Suzuki+etal:2008}.
  We implemented this method using MATLAB,
  which is a combination of the MIC code personally provided by 
  one of the authors,
  and the MLMI code available from the web page of one of the authors\footnote{
    \url{http://sugiyama-www.cs.titech.ac.jp/~sugi/software/MLMI/index.html}}.
  Following the suggestion provided in the original program code,
  MIC was initialized by pre-training of the kernel logistic model
  using the cluster assignments obtained by spectral clustering.
  The tuning parameter $\kernelparameter$ included in
  the sparse local-scaling kernel \eqref{sparse-local-scaling}
  was chosen from $\{1,\ldots,10\}$
  based on MLMI with Gaussian kernels (see Section~\ref{subsec:MIclustering}).
  The Gaussian kernel width in MLMI was chosen
  from $\{10^{-2},10^{-1.5},10^{-1},\ldots,10^2\}$
  based on cross-validation. 
  As suggested in the MLMI code provided by the author,
  the number of kernel bases in MLMI was limited to $200$,
  which were randomly chosen from all $\nsample$ kernels.

\item[SMIC:] SMI-based clustering with the sparse local-scaling kernel
  and the uniform class-prior distribution (see Section~\ref{subsec:SMIC}),
  where model selection is carried out by 
  least-squares MI \citep[LSMI;][see also Section~\ref{subsec:LSMI}]{BMCBio:Suzuki+etal:2009}.
  We implemented SMIC and LSMI using MATLAB by ourselves.
  The tuning parameter $\kernelparameter$ included in
  the sparse local-scaling kernel \eqref{sparse-local-scaling} 
  was chosen from $\{1,\ldots,10\}$
  based on LSMI with Gaussian kernels (see Section~\ref{subsec:LSMI}).
  The Gaussian kernel width and regularization parameter 
  included in LSMI were chosen from $\{10^{-2},10^{-1.5},10^{-1},\ldots,10^2\}$
  and $\{10^{-3},10^{-2.5},10^{-2},\ldots,10^1\}$, respectively,
  based on cross-validation. 
  Similarly to MLMI, the number of kernel bases in LSMI was limited to $200$,
  which were randomly chosen from all $\nsample$ kernels.
  
\end{description}


In addition to the clustering quality in terms of ARI,
we also evaluated the computational efficiency of each method 
by the CPU computation time.

\subsubsection{Datasets}
We used the following $6$ real-world datasets.

\begin{description}
\item[Digit $(d=256, n=5000,~\mathrm{and}~c=10)$:]
The \emph{USPS} hand-written digit dataset\footnote{
\url{http://www.gaussianprocess.org/gpml/data/}},
which contains $9298$ digit images.
Each image consists of $256$ $(=16 \times 16)$ pixels
and represents a digit in $\{0,1,2,\ldots,9\}$.
Each pixel takes a value in $[-1,+1]$
corresponding to the intensity level in gray-scale.
We randomly chose $500$ samples from each of the $10$ classes,
and used $5000$ samples in total.

\item[Face $(d=4096, n=100,~\mathrm{and}~c=10)$:]
The \emph{Olivetti Face} dataset\footnote{
\url{http://www.cs.toronto.edu/~roweis/data.html}},
which contains $400$ gray-scale face images
($40$ people; $10$ images per person).
Each image consists of $4096$ $(=64\times 64)$ pixels
and each pixel takes an integer value between $0$ and $255$ as the intensity level.
We randomly chose $10$ people, and used $100$ samples in total.

\item[Document $(d=50, n=700,~\mathrm{and}~c=7)$:]
The \emph{20-Newsgroups} dataset\footnote{
\url{http://people.csail.mit.edu/jrennie/20Newsgroups/}},
which contains $20000$ newsgroup documents
across $20$ different newsgroups.
We merged the $20$ newsgroups into the following $7$ top-level categories:
`\emph{comp}', `\emph{rec}', `\emph{sci}', `\emph{talk}', `\emph{alt}',
`\emph{misc}', and `\emph{soc}'.
Each document is expressed by a $10000$-dimensional \emph{bag-of-words} vector of 
\emph{term-frequencies}.
Following the convention \citep{Book:Joachims:2002},
we transformed the term-frequency vectors
to the \emph{term frequency/inverse document frequency} (TFIDF) vector,
i.e., we multiplied the term-frequency by
the logarithm of the inverse ratio of the documents containing the corresponding word.
We randomly chose $100$ samples from each of the $7$ classes,
and used $700$ samples in total.
We applied \emph{principal component analysis} \citep[PCA;][]{Pearson01,book:Jolliffe:1986}
to the $700$ samples, and extracted $50$-dimensional feature vectors.

\item[Word $(d=50, n=300,~\mathrm{and}~c=3)$:]
The \emph{SENSEVAL-2} dataset\footnote{
\url{http://www.senseval.org/}} for word-sense disambiguation.
We took the noun `\emph{interest}' appeared in $1930$ contexts,
having $3$ different meanings:
`advantage, advancement or favor',
`a share in a company or business',
and
`money paid for the use of money'
(i.e., $3$ classes).
From each surrounding context, we extracted a $14936$-dimensional
feature vector \citep{HLT-EMNLP:Niu+etal:2005},
which includes three types of features: 
\emph{part-of-speech} of neighboring words with position information,
\emph{bag-of-words} in the surrounding context, 
and \emph{local collocation} \citep{EMNLP:Lee+Ng:2002}.
We randomly chose $100$ samples from each of the $3$ classes,
and used $300$ samples in total.
We applied PCA to the $300$ samples, and 
extracted $50$-dimensional feature vectors.

\item[Accelerometry $(d=5, n=300,~\mathrm{and}~c=3)$:]
The \emph{ALKAN} dataset\footnote{
\url{http://alkan.mns.kyutech.ac.jp/web/data.html}},
which contains
$3$-axis (i.e., x-, y-, and z-axes) accelerometric data collected by
the \emph{iPod touch}.
In the data collection procedure,
subjects were asked to perform three specific tasks:
\emph{walking}, \emph{running}, and \emph{standing up}.
The duration of each task was arbitrary,
and the sampling rate was $20$Hz with small variations.
Each data-stream was then segmented in a sliding window manner
with window width $5$ seconds and sliding step $1$ second
\citep{NeuroComp:Hachiya+etal:2011}.
Depending on subjects, the position and orientation of
the accelerometer was arbitrary---held by hand or kept in a pocket or a bag.
For this reason, we took the $\ell_2$-norm of the $3$-dimensional acceleration vector
at each time step,
and computed the following $5$ orientation-invariant features from each window:
\emph{mean}, \emph{standard deviation},
\emph{fluctuation of amplitude}, \emph{average energy}, and \emph{frequency-domain entropy}
\citep{PerCom:Bao+Intille:2004,IFAWC:Bharatula+etal:2005}.
We randomly chose $100$ samples from each of the $3$ classes,
and used $300$ samples in total.

\item[Speech $(d=50, n=400,~\mathrm{and}~c=2)$:] An in-house speech dataset,
which contains short utterance samples recorded from
$2$ male subjects speaking in French with sampling rate $44.1$kHz.  
From each utterance sample, we extracted a $50$-dimensional
\emph{line spectral frequencies} vector \citep{ICASSP:Kain+Macon:1998}.
We randomly chose $200$ samples from each class,
and used $400$ samples in total.

\end{description}

For each dataset, the experiment was repeated $100$ times
with random choice of samples from the database,
where the cluster size is balanced.
Samples were centralized and their variance was normalized
in the dimension-wise manner, before feeding them to clustering algorithms.

\subsubsection{Results}

The experimental results are described in Table~\ref{tab:experimental-result}.
For the \emph{digit} dataset,
MIC and SMIC outperform KM, SC, and MNN in terms of ARI.
The entire computation time of SMIC including model selection
is faster than KM, SC, and MIC, and is comparable to MNN which
does not include a model selection procedure.
For the \emph{face} dataset,
SC, MIC, and SMIC are comparable to each other and are better than KM and MNN
in terms of ARI.
For the \emph{document} and \emph{word} datasets,
SMIC tends to outperform the other methods.
For the \emph{accelerometry} dataset,
MNN and SMIC work better than the other methods.
Finally, for the \emph{speech} dataset,
MIC and SMIC work comparably well, and are significantly better than KM, SC, and MNN.

\begin{table}[p]
  \centering
  \caption{Experimental results on real-world datasets (with equal cluster size).
    The average clustering accuracy (and its standard deviation in the bracket)
    in terms of ARI
    and the average CPU computation time in second
    over $100$ runs are described.
    Larger ARI is better, and shorter computation time is preferable.
    The best method in terms of the average ARI
    and methods judged to be comparable to the best one
    by the \emph{t-test} at the significance level $1\%$
    are described in boldface.
    Computation time of MIC and SMIC corresponds to
    the time for computing a clustering
    solution after model selection has been carried out.
    For references, computation time for the entire procedure
    including model selection is described
    in the square bracket, which depends on the number of model candidates
    (in the current setup, we had $81$ ($=9\times9$) candidates.
}
  \label{tab:experimental-result}
  \begin{tabular}{@{}c|cccccc@{}}
\multicolumn{6}{c}{}\\ 
\multicolumn{6}{c}{Digit ($d=256,~n=5000,  ~\mathrm{and}~c=10$)}\\ 
&
KM &
SC &
MNN &
MIC &
SMIC \\
\hline
ARI & 
        0.42(0.01)  & 
        0.24(0.02)  & 
        0.44(0.03)  & 
\textbf{0.63(0.08)} & 
\textbf{0.63(0.05)} \\ 
Time & 
        835.9  & 
        973.3  & 
        318.5  & 
        84.4[3631.7]  & 
        14.4[359.5]  \\ 
\multicolumn{6}{c}{}\\ 
\multicolumn{6}{c}{Face ($d=4096,~n=100,  ~\mathrm{and}~c=10$)}\\ 
&
KM &
SC &
MNN &
MIC &
SMIC \\
\hline
ARI & 
        0.60(0.11)  & 
\textbf{0.62(0.11)} & 
        0.47(0.10)  & 
\textbf{0.64(0.12)} & 
\textbf{0.65(0.11)} \\ 
Time & 
        93.3  & 
        2.1  & 
        1.0  & 
        1.4[30.8]  & 
        0.0[19.3]  \\ 
\multicolumn{6}{c}{}\\ 
\multicolumn{6}{c}{Document ($d=50,~ n=700,  ~\mathrm{and}~c=7$)}\\ 
&
KM &
SC &
MNN &
MIC &
SMIC \\
\hline
ARI & 
        0.00(0.00)  & 
        0.09(0.02)  & 
        0.09(0.02)  & 
        0.01(0.02)  & 
\textbf{0.19(0.03)} \\ 
Time & 
        77.8  & 
        9.7  & 
        6.4  & 
        3.4[530.5]  & 
        0.3[115.3]  \\ 
\multicolumn{6}{c}{}\\ 
\multicolumn{6}{c}{Word ($d=50,~n=300,  ~\mathrm{and}~c=3$)}\\ 
&
KM &
SC &
MNN &
MIC &
SMIC \\
\hline
ARI & 
        0.04(0.05)  & 
        0.02(0.01)  & 
        0.02(0.02)  & 
        0.04(0.04)  & 
\textbf{0.08(0.05)} \\ 
Time & 
        6.5  & 
        5.9  & 
        2.2  & 
        1.0[369.6]  & 
        0.2[203.9]  \\ 
\multicolumn{6}{c}{}\\ 
\multicolumn{6}{c}{Accelerometry ($d=5,~n=300,  ~\mathrm{and}~c=3$)}\\ 
&
KM &
SC &
MNN &
MIC &
SMIC \\
\hline
ARI & 
        0.49(0.04)  & 
        0.58(0.14)  & 
\textbf{0.71(0.05)} & 
        0.57(0.23)  & 
\textbf{0.68(0.12)} \\ 
Time & 
        0.4  & 
        3.3  & 
        1.9  & 
        0.8[410.6]  & 
        0.2[92.6]  \\ 
\multicolumn{6}{c}{}\\ 
\multicolumn{6}{c}{Speech ($d=50,~n=400,  ~\mathrm{and}~c=2$)}\\ 
&
KM &
SC &
MNN &
MIC &
SMIC \\
\hline
ARI & 
        0.00(0.00)  & 
        0.00(0.00)  & 
        0.04(0.15)  & 
\textbf{0.18(0.16)} & 
\textbf{0.21(0.25)} \\ 
Time & 
        0.9  & 
        4.2  & 
        1.8  & 
        0.7[413.4]  & 
        0.3[179.7]  \\ 
  \end{tabular}
\end{table}  

\begin{table}[t]
  \centering
  \caption{Experimental results on real-world datasets under imbalanced setup.
    ARI values are described in the table.
    Class-imbalance was realized by setting the sample size
    of the first class $m$ times larger than other classes.
    SMIC was computed with the uniform prior (i.e., the non-informative prior).
    The results for $m=1$ are the same as the ones reported in 
    Table~\ref{tab:experimental-result}.
}
  \label{tab:experimental-result2}
  \begin{tabular}{@{}c|cccccc@{}}
\multicolumn{6}{c}{}\\ 
\multicolumn{6}{c}{Digit ($d=256,~n=5000,  ~\mathrm{and}~c=10$)}\\ 
&
KM &
SC &
MNN &
MIC &
SMIC \\
\hline
$m=1$ & 
        0.42(0.01)  & 
        0.24(0.02)  & 
        0.44(0.03)  & 
\textbf{0.63(0.08)} & 
\textbf{0.63(0.05)} \\ 
$m=2$ & 
        0.52(0.01)  & 
        0.21(0.02)  & 
        0.43(0.04)  & 
        0.60(0.05)  & 
\textbf{0.63(0.05)} \\ 
\multicolumn{6}{c}{}\\ 
\multicolumn{6}{c}{Document ($d=50,~ n=700,  ~\mathrm{and}~c=7$)}\\ 
&
KM &
SC &
MNN &
MIC &
SMIC \\
\hline
$m=1$ & 
        0.00(0.00)  & 
        0.09(0.02)  & 
        0.09(0.02)  & 
        0.01(0.02)  & 
\textbf{0.19(0.03)} \\ 
$m=2$ & 
        0.01(0.01)  & 
        0.10(0.03)  & 
        0.10(0.02)  & 
        0.01(0.02)  & 
\textbf{0.19(0.04)} \\ 
$m=3$ & 
        0.01(0.01)  & 
        0.10(0.03)  & 
        0.09(0.02)  & 
        -0.01(0.03)  & 
\textbf{0.16(0.05)} \\ 
$m=4$ & 
        0.02(0.01)  & 
        0.09(0.03)  & 
        0.08(0.02)  & 
        -0.00(0.04)  & 
\textbf{0.14(0.05)} \\ 
\multicolumn{6}{c}{}\\ 
\multicolumn{6}{c}{Word ($d=50,~n=300,  ~\mathrm{and}~c=3$)}\\ 
&
KM &
SC &
MNN &
MIC &
SMIC \\
\hline
$m=1$ & 
        0.04(0.05)  & 
        0.02(0.01)  & 
        0.02(0.02)  & 
        0.04(0.04)  & 
\textbf{0.08(0.05)} \\ 
$m=2$ & 
        0.00(0.07)  & 
        -0.01(0.01)  & 
        0.01(0.02)  & 
        -0.02(0.05)  & 
\textbf{0.03(0.05)} \\ 
\multicolumn{6}{c}{}\\ 
\multicolumn{6}{c}{Accelerometry ($d=5,~n=300,  ~\mathrm{and}~c=3$)}\\ 
&
KM &
SC &
MNN &
MIC &
SMIC \\
\hline
$m=1$ & 
        0.49(0.04)  & 
        0.58(0.14)  & 
\textbf{0.71(0.05)} & 
        0.57(0.23)  & 
\textbf{0.68(0.12)} \\ 
$m=2$ & 
        0.48(0.05)  & 
        0.54(0.14)  & 
        0.58(0.11)  & 
        0.49(0.19)  & 
\textbf{0.69(0.16)} \\ 
$m=3$ & 
        0.49(0.05)  & 
        0.47(0.10)  & 
        0.42(0.12)  & 
        0.42(0.14)  & 
\textbf{0.66(0.20)} \\ 
$m=4$ & 
        0.49(0.06)  & 
        0.38(0.11)  & 
        0.31(0.09)  & 
        0.40(0.18)  & 
\textbf{0.56(0.22)} \\ 
  \end{tabular}
\end{table}

Overall, MIC was shown to work reasonably well,
implying that the MLMI-based model selection strategy is practically useful.
SMIC was shown to work even better than MIC, with much less computation time.
The accuracy improvement of SMIC over MIC was gained by computing the SMIC solution in
a closed-form without any heuristic initialization.
The computational efficiency of SMIC was brought by
the analytic computation of the optimal solution and the class-wise optimization of LSMI
(see Section~\ref{subsec:LSMI}).

The performance of MNN and SC was rather unstable because of
the heuristic averaging of the number of nearest neighbors in MNN
and the heuristic choice of local scaling in SC.
In terms of computation time, they are relatively efficient
for small- to medium-sized datasets,
but they are expensive for the largest dataset, \emph{digit}.
KM was not reliable for the \emph{document} and \emph{speech} datasets
because of the restriction that the cluster boundaries are linear.
For the \emph{digit}, \emph{face}, and \emph{document} datasets,
KM was computationally very expensive
since a large number of iterations were needed
until convergence to a local optimum solution.

Finally, we performed similar experiments under imbalanced setup,
where the sample size
of the first class was set to be $m$ times larger than other classes
with the total number of samples fixed to the same number\footnote{
Because of the dataset size,
this experiment was carried out only for several cases.
See Table~\ref{tab:experimental-result2}.}.
The results are summarized in Table~\ref{tab:experimental-result2},
showing that the performance of all methods
tends to be degraded as the degree of cluster imbalance increases.
Thus, clustering becomes more challenging if the cluster size is imbalanced.
Among the compared methods, the proposed SMIC (with the uniform prior)
still worked better than other methods.

Overall, the proposed SMIC combined with LSMI was shown to be
a useful alternative to existing clustering approaches.
 

\section{Conclusions}
\label{sec:conclusion}
In this paper, we proposed a novel \emph{information-maximization clustering} method
that learns class-posterior probabilities
in an unsupervised manner so that the \emph{squared-loss mutual information} (SMI)
between feature vectors and cluster assignments is maximized.
The proposed algorithm, called \emph{SMI-based clustering} (SMIC),
allows us to obtain clustering solutions \emph{analytically}
by solving a kernel eigenvalue problem.
Thus, unlike the previous information-maximization
clustering methods \citep{NIPS2005_569,NIPS2010_0457},
SMIC does not suffer from the problem of local optima.
Furthermore, we proposed to use an optimal non-parametric
SMI estimator called \emph{least-squares mutual information} (LSMI)
for data-driven parameter optimization.
Through experiments, SMIC combined with LSMI was demonstrated to compare favorably
with existing clustering methods.

In experiments, the proposed clustering method was shown to be useful
for various types of data.
However, the amount of improvement is large for some datasets,
while it is mild for other datasets.
It is thus practically important to have more insights on
in what case the proposed method is advantageous.

The sparse local-scaling kernel \eqref{sparse-local-scaling} was shown to be
useful in experiments.
Since this produces a sparse kernel matrix, the computation of SMIC
(i.e., solving a kernel eigenvalue problem) can be carried out very efficiently.
However, if model selection is taken into account,
the proposed clustering procedure is still computationally
rather demanding due to the repeated computation of LSMI,
which requires to solve a system of linear equations.
In the experiments,
we used the Gaussian kernel \eqref{Gaussian-kernel} for LSMI 
and found it useful in practice.
However, it produces a dense kernel matrix and
thus a dense system of linear equations need to be solved,
which is computationally expensive.
If a sparse kernel is used also for LSMI,
its computational efficiency will be highly improved.
In our preliminary experiments,
the use of the sparse local-scaling kernel for LSMI
improved the computational efficiency,
but it did not perform as well as the Gaussian kernel.
Thus, our important future work is to find a sparse kernel
that gives an accurate approximation of SMI with high computational efficiency.

As addressed in \citet{ICML:Song+etal:2007b}, kernelized methods 
can be applied to clustering of \emph{non-vectorial structured objects}
such as \emph{strings}, \emph{trees}, and \emph{graphs}
by employing kernel functions defined for such structured data
\citep{JMLR:Lodhi+etal:2002,nips02-AA58,ICML:Kashima+Koyanagi:2002,%
ICML:Kondor+Lafferty:2002,ICML:Kashima+etal:2003,COLT:Gaertner+etal:2003,SIGKDD-ex:Gaertner:2003}.
Since these structured kernels usually contain tuning parameters,
the performance of clustering methods without systematic model selection strategies
depends on subjective parameter tuning,
which is not preferable in practice.
For Gaussian kernels, there exists a popular heuristic 
that the Gaussian width is set to the median distance
between samples \citep{AS:Fukumizu+etal:2009}.
However, there seems no such common heuristic for structured kernels.
In such scenarios, the proposed method will be highly advantageous
because it allows systematic model selection for any kernels.
We will explore this direction in our future work.

We experimentally showed
that the proposed method with the uniform class-prior
distribution still works reasonably well even when the true class-prior
probability is not uniform.
This is a useful property in practice since 
the true class-prior probability is often unknown.
Another way to address this issue is to estimate
the true class-prior probability in a data-driven fashion,
for example, iteratively performing clustering and updating the
class-prior probabilities. We will investigate such an adaptive approach
in our future work.

The proposed method uses SMI as the common guidance for clustering,
although we are using two SMI approximators:
$\widehat{\mathrm{SMI}}$ defined by Eq.\eqref{SMIhat}
for finding clustering solutions
and $\mathrm{LSMI}$ defined by Eq.\eqref{LSMI}
for selecting models.
Since $\widehat{\mathrm{SMI}}$ does not explicitly include cluster labels $\{y_i\}_{i=1}^\nsample$,
it has a simple form and therefore is suited for efficient maximization.
Indeed, we can obtain an optimal solution analytically by 
solving an eigenvalue problem.
However, since $\widehat{\mathrm{SMI}}$ is an unsupervised estimator
where the cluster labels $\{y_i\}_{i=1}^\nsample$ are not used,
it may not be accurate enough for model selection purposes.
Indeed, our preliminary experiments showed that
the use of $\widehat{\mathrm{SMI}}$ is not appropriate as a model selection criterion.
On the other hand, since $\mathrm{LSMI}$ achieves the optimal non-parametric convergence rate,
its high accuracy is suitable for model selection purposes.
However, LSMI explicitly requires cluster labels $\{y_i\}_{i=1}^\nsample$
and thus is not suited for efficient maximization.
Based on the optimality of LSMI, we ideally want to use LSMI consistently
for \emph{both} finding clustering solutions and selecting models.
However, its optimization involves discrete optimization of
$\{y_i\}_{i=1}^\nsample$, which is cumbersome in practice.
Our future challenge is to develop a practical clustering algorithm
based directly on LSMI.





\section*{Acknowledgments}
We would like to thank Ryan Gomes
for providing us his program code of information-maximization clustering.
MS was supported by
SCAT, AOARD, and the FIRST program.
MY and MK were supported by the JST PRESTO program,
and HH was supported by the FIRST program.

\appendix
\section*{Appendix: Rand Index and Adjusted Rand Index}
Here, we review the definitions
of the \emph{Rand index} \citep[RI;][]{JASA:Rand:1971}
and the \emph{adjusted Rand index} \citep[ARI;][]{JoC:Hubert+Arabie:1985},
which are used for evaluating the quality of clustering results.
Let $\{y_i^*\}_{i=1}^\nsample$ be the ground-truth cluster assignments,
and let $\{y_i\}_{i=1}^\nsample$ be a clustering solution obtained
by some algorithm.
The goal is to quantitatively evaluate the similarity
between $\{y_i\}_{i=1}^\nsample$ and $\{y_i^*\}_{i=1}^\nsample$.

\begin{table}[t]
  \centering
  \caption{Notation for Rand index and adjusted Rand index.}
  \label{tab:ARI}
  \vspace*{2mm}
  \begin{tabular}{c@{~~~~~~~~~~~}c}
  (a)&(b)\\[2mm]
  \begin{tabular}{c|ccc|c}
    & $\calC_1^*$ & $\cdots$ & $\calC_{\nclass}^*$ & Sum\\
    \hline
    $\calC_1$ & $\nsample_{1,1}$ & $\cdots$ & $\nsample_{1,\nclass}$ & $\nsample_1$\\
    $\vdots$ & $\vdots$ & $\ddots$ & $\vdots$ & $\vdots$ \\
    $\calC_{\nclass}$ & $\nsample_{\nclass,1}$ & $\cdots$ & $\nsample_{\nclass,\nclass}$ & $\nsample_{\nclass}$\\
    \hline
    Sum & $\nsample_1^*$ & $\cdots$ & $\nsample_{\nclass}^*$ & $\nsample$\\
  \end{tabular}
&
  \begin{tabular}{c|c|c|c}
\multicolumn{2}{c|}{}&\multicolumn{2}{c}{Pairs in $\{\calC_{y'}^*\}_{y'=1}^{\nclass}$}\\
\cline{3-4}
\multicolumn{2}{c|}{}& Same & Different\\
\hline
Pairs in
& Same & $m_{\calC,\calC^*}$ & $m_{\calC,\bar{\calC}^*}$\\
\cline{2-4}
$\{\calC_y\}_{y=1}^{\nclass}$
& Different & $m_{\bar{\calC},\calC^*}$ & $m_{\bar{\calC},\bar{\calC}^*}$ \\
  \end{tabular}
\end{tabular}
\end{table}

The most direct way to evaluate the discrepancy
between $\{y_i\}_{i=1}^\nsample$ and $\{y_i^*\}_{i=1}^\nsample$
would be to naively verify the correctness of the predicted labels.
However, in clustering, 
predicted class labels $\{y_i\}_{i=1}^\nsample$ do not have to be
equal to the true labels $\{y_i^*\}_{i=1}^\nsample$,
but only their \emph{partition} matters.
The correctness of the partition may be evaluated
by verifying the correctness of the predicted labels for all possible label permutations.
However, this is computationally expensive if the number of classes is large.
RI and ARI are alternative performance measures that can overcome this computational problem
in a systematic way.

For the two partitions $\{y_i\}_{i=1}^\nsample$ and $\{y_i^*\}_{i=1}^\nsample$,
let $\calC_y$ and $\calC_y^*$ ($y=1,\ldots,\nclass$)
be sets of indices of samples in cluster $y$, respectively:
\begin{align*}
  \calC_y&=\{y_i\;|\;y_i=y\},\\
  \calC_y^*&=\{y_i^*\;|\;y_i^*=y\}.
\end{align*}
Let $\nsample_{y,y'}$ be the number of samples that are assigned to
the cluster $\calC_y$ and the cluster $\calC_{y'}^*$.
Let $\nsample_y$ (resp.~$\nsample_y^*$) be
the number of samples that are assigned to the cluster $\calC_y$ (resp.~$\calC_{y'}^*$).
The notation is summarized in Table~\ref{tab:ARI}(a).

Let $m_{\calC,\calC^*}$, $m_{\calC,\bar{\calC}^*}$, $m_{\bar{\calC},\calC^*}$, and $m_{\bar{\calC},\bar{\calC}^*}$
be defined as
\begin{align*}
  m_{\calC,\calC^*}&:=\sum_{y,y'=1}^{\nclass}
  \begin{pmatrix}
    \nsample_{y,y'}\\ 2\\
  \end{pmatrix},\\
  m_{\calC,\bar{\calC}^*}&:=
  \sum_{y=1}^{\nclass}
  \begin{pmatrix}
    \nsample_{y}\\ 2\\
  \end{pmatrix}
  -m_{\calC,\calC^*},\\
  m_{\bar{\calC},\calC^*}&:=
  \sum_{y'=1}^{\nclass}
  \begin{pmatrix}
    \nsample_{y'}^*\\ 2\\
  \end{pmatrix}
  -m_{\calC,\calC^*},\\
  m_{\bar{\calC},\bar{\calC}^*}&:=
  \begin{pmatrix}
    \nsample\\ 2\\
  \end{pmatrix}
  -m_{\calC,\calC^*}-m_{\calC,\bar{\calC}^*}-m_{\bar{\calC},\calC^*},
\end{align*}
where $m_{\calC,\calC^*}$ denotes the number of pairs of samples that are assigned to the same cluster
both in $\{\calC_y\}_{y=1}^{\nclass}$ and $\{\calC_{y'}^*\}_{y'=1}^{\nclass}$,
$m_{\calC,\bar{\calC}^*}$ denotes the number of pairs of samples that are assigned to the same cluster
in $\{\calC_y\}_{y=1}^{\nclass}$ but are assigned to different clusters in $\{\calC_{y'}^*\}_{y'=1}^{\nclass}$,
$m_{\bar{\calC},\calC^*}$ denotes the number of pairs of samples that are assigned to the same cluster
in $\{\calC_{y'}^*\}_{y'=1}^{\nclass}$ but are assigned to different clusters in $\{\calC_y\}_{y=1}^{\nclass}$,
and
$m_{\bar{\calC},\bar{\calC}^*}$ denotes the number of pairs of samples that are assigned to
different clusters both in $\{\calC_y\}_{y=1}^{\nclass}$ and $\{\calC_{y'}^*\}_{y'=1}^{\nclass}$.
$m_{\calC,\calC^*}+m_{\bar{\calC},\bar{\calC}^*}$ can be considered as
the number of `agreements' between $\{\calC_y\}_{y=1}^{\nclass}$ and $\{\calC_{y'}^*\}_{y'=1}^{\nclass}$,
while $m_{\calC,\bar{\calC}^*}+m_{\bar{\calC},\calC^*}$ can be regarded as
the number of `disagreements' between $\{\calC_y\}_{y=1}^{\nclass}$ and $\{\calC_{y'}^*\}_{y'=1}^{\nclass}$.
The notation is summarized in Table~\ref{tab:ARI}(b).

The \emph{Rand index} \citep[RI;][]{JASA:Rand:1971} is defined and expressed as
\begin{align*}
  \mathrm{RI}&:=\frac{m_{\calC,\calC^*}+m_{\bar{\calC},\bar{\calC}^*}}
  {m_{\calC,\calC^*}+m_{\calC,\bar{\calC}^*}+m_{\bar{\calC},\calC^*}+m_{\bar{\calC},\bar{\calC}^*}}\\
  &\phantom{:}=
  (m_{\calC,\calC^*}+m_{\bar{\calC},\bar{\calC}^*})\Big/
  \begin{pmatrix}
    \nsample\\ 2\\
  \end{pmatrix}.
\end{align*}
The Rand index lies between $0$ and $1$, and takes $1$ if the two clustering solutions
$\{\calC_y\}_{y=1}^{\nclass}$ and $\{\calC_{y'}^*\}_{y'=1}^{\nclass}$
agree with each other perfectly.

A potential drawback of the Rand index is that
its expected value is not a constant (say, $0$) 
if two clustering solutions are completely random.
To overcome this problem,
the \emph{adjusted Rand index} (ARI) was proposed
\citep{JoC:Hubert+Arabie:1985}.
ARI is defined as
\begin{align*}
   \mathrm{ARI}:=\frac{m_{\calC,\calC^*}+m_{\bar{\calC},\bar{\calC}^*}-\mu}
   {m_{\calC,\calC^*}+m_{\calC,\bar{\calC}^*}+m_{\bar{\calC},\calC^*}+m_{\bar{\calC},\bar{\calC}^*}-\mu}.
\end{align*}
$\mu$ is the expected value of $m_{\calC,\calC^*}+m_{\bar{\calC},\bar{\calC}^*}$:
\begin{align*}
  \mu:=\mathbbE\left[m_{\calC,\calC^*}+m_{\bar{\calC},\bar{\calC}^*}\right],
\end{align*}
where $\mathbbE$ denotes the expectation over cluster assignments.
ARI takes the maximum value $1$ when two sets of cluster assignments are identical,
and takes $0$ if the index equals its expected value.

Under the assumption that
the clustering solutions $\{\calC_y\}_{y=1}^{\nclass}$ and $\{\calC_{y'}^*\}_{y'=1}^{\nclass}$
are randomly drawn from a generalized hyper-geometric distribution,
it holds that
\begin{align*}
  \mathbbE\left[m_{\calC,\calC^*}\right]
    &=(m_{\calC,\calC^*}+m_{\calC,\bar{\calC}^*})(m_{\calC,\calC^*}+m_{\bar{\calC},\calC^*})
    \Big/
  \begin{pmatrix}
    \nsample\\ 2\\
  \end{pmatrix},\\
  \mathbbE\left[m_{\bar{\calC},\bar{\calC}^*}\right]
    &=(m_{\calC,\bar{\calC}^*}+m_{\bar{\calC},\bar{\calC}^*})(m_{\bar{\calC},\calC^*}+m_{\bar{\calC},\bar{\calC}^*})
    \Big/
  \begin{pmatrix}
    \nsample\\ 2\\
  \end{pmatrix}.
\end{align*}
Then ARI can be expressed as
\begin{align*}
     \mathrm{ARI}=
     \frac{\displaystyle
  \begin{pmatrix}
    \nsample\\ 2\\
  \end{pmatrix}
\sum_{y,y'=1}^{\nclass}
  \begin{pmatrix}
    \nsample_{y,y'}\\ 2\\
  \end{pmatrix}
-
  \sum_{y=1}^{\nclass}
  \begin{pmatrix}
    \nsample_{y}\\ 2\\
  \end{pmatrix}
  \sum_{y'=1}^{\nclass}
  \begin{pmatrix}
    \nsample_{y'}^*\\ 2\\
  \end{pmatrix}
}{\displaystyle
\frac{1}{2}
  \begin{pmatrix}
    \nsample\\ 2\\
  \end{pmatrix}
\left[
  \sum_{y=1}^{\nclass}
  \begin{pmatrix}
    \nsample_{y}\\ 2\\
  \end{pmatrix}
+
  \sum_{y'=1}^{\nclass}
  \begin{pmatrix}
    \nsample_{y'}^*\\ 2\\
  \end{pmatrix}
\right]
-
  \sum_{y=1}^{\nclass}
  \begin{pmatrix}
    \nsample_{y}\\ 2\\
  \end{pmatrix}
  \sum_{y'=1}^{\nclass}
  \begin{pmatrix}
    \nsample_{y'}^*\\ 2\\
  \end{pmatrix}
}.
\end{align*}

Note that RI and ARI can be defined even when
two sets of cluster assignments $\{y_i\}_{i=1}^\nsample$ and $\{y_i^*\}_{i=1}^\nsample$
have different numbers of clusters,
i.e., $\{\calC_y\}_{y=1}^{\nclass}$ and $\{\calC_{y'}^*\}_{y'=1}^{\nclass'}$
with $\nclass\neq\nclass'$.
This is highly convenient in practice since, when the number of true clusters is large,
clustering algorithms often produce clustering solutions with a smaller number
of clusters (i.e., some of the clusters have no samples).
Even in such cases,
RI and ARI can still be used for evaluating the quality of clustering solutions.

\bibliography{E:/bib/sugiyama,E:/bib/mybib}

\end{document}

%% file: mathdef.tex
\newcommand{\argmin}{\mathop{\mathrm{argmin\,}}}
\newcommand{\argmax}{\mathop{\mathrm{argmax\,}}}

\newcommand{\tr}[1]{\mathrm{tr}(#1)}

\newcommand{\inner}[2]{\langle #1,#2\rangle}

\newcommand{\mathbbE}{\mathbb{E}}

\newcommand{\mathbbR}{\mathbb{R}}

\newcommand{\boldzero}{{\boldsymbol{0}}}
\newcommand{\boldone}{{\boldsymbol{1}}}

\newcommand{\boldA}{{\boldsymbol{A}}}

\newcommand{\boldD}{{\boldsymbol{D}}}

\newcommand{\boldH}{{\boldsymbol{H}}}
\newcommand{\boldI}{{\boldsymbol{I}}}

\newcommand{\boldK}{{\boldsymbol{K}}}

\newcommand{\boldP}{{\boldsymbol{P}}}

\newcommand{\boldW}{{\boldsymbol{W}}}
\newcommand{\boldX}{{\boldsymbol{X}}}

\newcommand{\boldc}{{\boldsymbol{c}}}

\newcommand{\boldh}{{\boldsymbol{h}}}

\newcommand{\boldx}{{\boldsymbol{x}}}
\newcommand{\boldy}{{\boldsymbol{y}}}

\newcommand{\boldalpha}{{\boldsymbol{\alpha}}}
\newcommand{\boldbeta}{{\boldsymbol{\beta}}}

\newcommand{\boldtheta}{{\boldsymbol{\theta}}}

\newcommand{\boldlambda}{{\boldsymbol{\lambda}}}
\newcommand{\boldmu}{{\boldsymbol{\mu}}}

\newcommand{\boldxi}{{\boldsymbol{\xi}}}
\newcommand{\boldpi}{{\boldsymbol{\pi}}}

\newcommand{\boldphi}{{\boldsymbol{\phi}}}

\newcommand{\boldGamma}{{\boldsymbol{\Gamma}}}

\newcommand{\boldPi}{{\boldsymbol{\Pi}}}

\newcommand{\calC}{{\mathcal{C}}}

\newcommand{\calN}{{\mathcal{N}}}

\newcommand{\calX}{{\mathcal{X}}}
\newcommand{\calY}{{\mathcal{Y}}}
\newcommand{\calZ}{{\mathcal{Z}}}

\newcommand{\thetah}{{\widehat{\theta}}}

\newcommand{\boldthetah}{{\widehat{\boldtheta}}}

\newcommand{\kernelparameter}{t}
\newcommand{\Kernelparameter}{T}
\newcommand{\kernelparameterr}{\gamma}
\newcommand{\Kernelparameterr}{\Gamma}
\newcommand{\regularizationparameterr}{\delta}
\newcommand{\Regularizationparameterr}{\Delta}

\newcommand{\inputdim}{d}

\newcommand{\ratiosymbol}{r}
 \newcommand{\ratio}{\ratiosymbol^\ast}
 \newcommand{\ratioh}{\widehat{\ratiosymbol}}
 \newcommand{\ratiomodel}{\ratiosymbol}

\newcommand{\densitysymbol}{p}

\newcommand{\density}{\densitysymbol^\ast}

\newcommand{\densityh}{\widehat{\densitysymbol}}

\newcommand{\densitymodel}{\densitysymbol}
\newcommand{\densitymodell}{\densitysymbol}

\newcommand{\nsample}{n}

\newcommand{\nclass}{c}

\newcommand{\boldHh}{{\widehat{\boldH}}}
\newcommand{\boldhh}{{\widehat{\boldh}}}

\newcommand{\Hh}{{\widehat{H}}}
\newcommand{\hh}{{\widehat{h}}}

\newcommand{\boldalphah}{\widehat{\boldalpha}}

